\documentclass[10pt,journal,compsoc]{IEEEtran}
\usepackage[switch]{lineno}
\usepackage{lipsum}

\ifCLASSOPTIONcompsoc
  \usepackage[nocompress]{cite}
\else
  \usepackage{cite}
\fi
\ifCLASSINFOpdf
\else
\fi
\usepackage{amsmath}
\usepackage{graphicx}
\usepackage{multirow}
\usepackage{amsfonts}
\usepackage{xcolor}
\usepackage{array}

\definecolor{newgreen}{rgb}{0.05, 0.5, 0.06}
\definecolor{newblue}{rgb}{0.0, 0.47, 0.75}
\definecolor{newyellow}{rgb}{0.98, 0.85, 0.37}
\definecolor{newred}{rgb}{0.89, 0.0, 0.13}

\newcommand{\re}[1]{\textcolor{black}{#1}}

\usepackage{url}
\usepackage{xcolor}
\usepackage{makecell}

\begin{document}

% \linenumbers

\title{Saying the Unseen:\\ Video Descriptions via Dialog Agents}

\author{Ye~Zhu,~
        Yu~Wu,~
        Yi~Yang,~
        and~Yan~Yan$^{\star}$~% <-this % stops a space
\IEEEcompsocitemizethanks{\IEEEcompsocthanksitem Ye Zhu and Yan Yan are with the Department of Computer Science, Illinois Institute of Technology, USA. (Corresponding Author: Yan Yan) \protect\\
E-mail: yzhu96@hawk.iit.edu, yyan34@iit.edu
%<-this % stops an unwanted space
\IEEEcompsocthanksitem Yu Wu and Yi Yang are with University of Technology Sydney, Australia. \protect\\
E-mail: yu.wu-3@student.uts.edu.au,~yi.yang@uts.edu.au 
\IEEEcompsocthanksitem Codes will be updated at: https://github.com/L-YeZhu/Video-Description-via-Dialog-Agents-ECCV2020
}
% \thanks{Manuscript received April 19, 2005; revised August 26, 2015.}
}


\markboth{IEEE Transactions on Pattern Analysis and Machine Intelligence, 2021}%
{Y. Zhu \MakeLowercase{\textit{et al.}}:}



\IEEEtitleabstractindextext{%
\begin{abstract}
\sloppy
% Creating secure and reliable Artificial Intelligence (AI) systems with interpretability is the common pursuit of researchers in the machine learning community. 
Current vision and language tasks usually take complete visual data (\textit{e.g.}, raw images or videos) as input, however, practical scenarios may often consist the situations where part of the visual information becomes inaccessible due to various reasons {\textit{e.g.}, restricted view with fixed camera or intentional vision block for security concerns}.
As a step towards the more practical application scenarios, we introduce a novel task that aims to describe a video using the natural language dialog between two agents as a supplementary information source given incomplete visual data.
Different from most existing vision-language tasks where AI systems have full access to images or video clips, which may reveal sensitive information such as recognizable human faces or voices, we \re{intentionally limit the visual input for AI systems and seek a more secure and transparent information medium, \textit{i.e.}, the natural language dialog, to supplement the missing visual information.}
Specifically, one of the intelligent agents - \emph{Q-BOT} - is given two semantic segmented frames from the beginning and the end of the video, as well as a finite number of opportunities to ask relevant natural language questions before describing the unseen video. \emph{A-BOT}, the other agent who has access to the entire video, assists \emph{Q-BOT} to accomplish the goal by answering the asked questions.
We introduce two different experimental settings with either a generative (\textit{i.e.}, agents generate questions and answers freely) or a discriminative (\textit{i.e.}, agents select the questions and answers from candidates) internal dialog generation process. With the proposed unified QA-Cooperative networks, we experimentally demonstrate the knowledge transfer process between the two dialog agents and the effectiveness of using the natural language dialog as a supplement for incomplete implicit visions.

\end{abstract}

\begin{IEEEkeywords}
Video Description, Dialog Agents, Multi-modal Learning.
\end{IEEEkeywords}}
\maketitle

\IEEEdisplaynontitleabstractindextext

% \fi
\IEEEpeerreviewmaketitle

\IEEEraisesectionheading{\section{Introduction}\label{sec:introduction}}

% for IEEE Computer Society journal papers produced under \LaTeX\ using
% IEEEtran.cls version 1.8b and later.
% % You must have at least 2 lines in the paragraph with the drop letter
% % (should never be an issue)
% I wish you the best of success.

 

\IEEEPARstart{C}{lassic} vision-language tasks such as \re{video} captioning and visual question answering (VQA) have been well explored in previous work\re{~\cite{antol2015vqa,das2017visual,de2017guesswhat,alamri2019audio,song2018explore,zhou2018end,wang2018bidirectional,wang2018reconstruction}} and have achieved promising performance. 
Most existing research studies on these tasks provide AI systems with full access to images or videos.
However, these images or videos may reveal sensitive personal biometric information (\textit{e.g.}, recognizable human faces or voices), thus aggravating the arising concerns on the privacy and security issues of AI from the general public in recent years.
Although directly taking the original \re{visual} data as input usually helps with the performance improvement, we also observe that it is not always necessary to fulfill the final task in practical scenarios (\textit{e.g.}, we do not need to directly look at the human faces to tell their actions or gestures). 
In addition, in more practical application scenarios, we may encounter the situations where part of the visual information is inaccessible due to reasons such as restricted view of fixed cameras.
Based on the above observations, we make efforts to introduce a new multi-agent task that aims to describe a video based on implicit visions in this work. The concept of implicit vision refers to the idea that the given visual information is intentionally made incomplete to protect user privacy. We then propose to supplement the missing visual information via a less sensitive information medium, \textit{i.e.}, the natural language dialog.
Unlike video clips, AI systems, or even humans, can hardly identify the biometric information of a person based on the natural language descriptions from the dialog.
In addition, natural language dialog is more transparent for humans in the sense that humans can understand and interpret the sentences compared to traditional obscure feature embedding in matrix forms. 
Overall, we have two objectives to fulfill in this work: to propose a novel video description task setup that addresses the privacy concerns by providing implicit visual data, and to demonstrate that the natural language dialog can be a more secure yet effective source to supplement the missing visual information.

\begin{figure*}[t]
    \centering
    \includegraphics[width=0.75\textwidth]{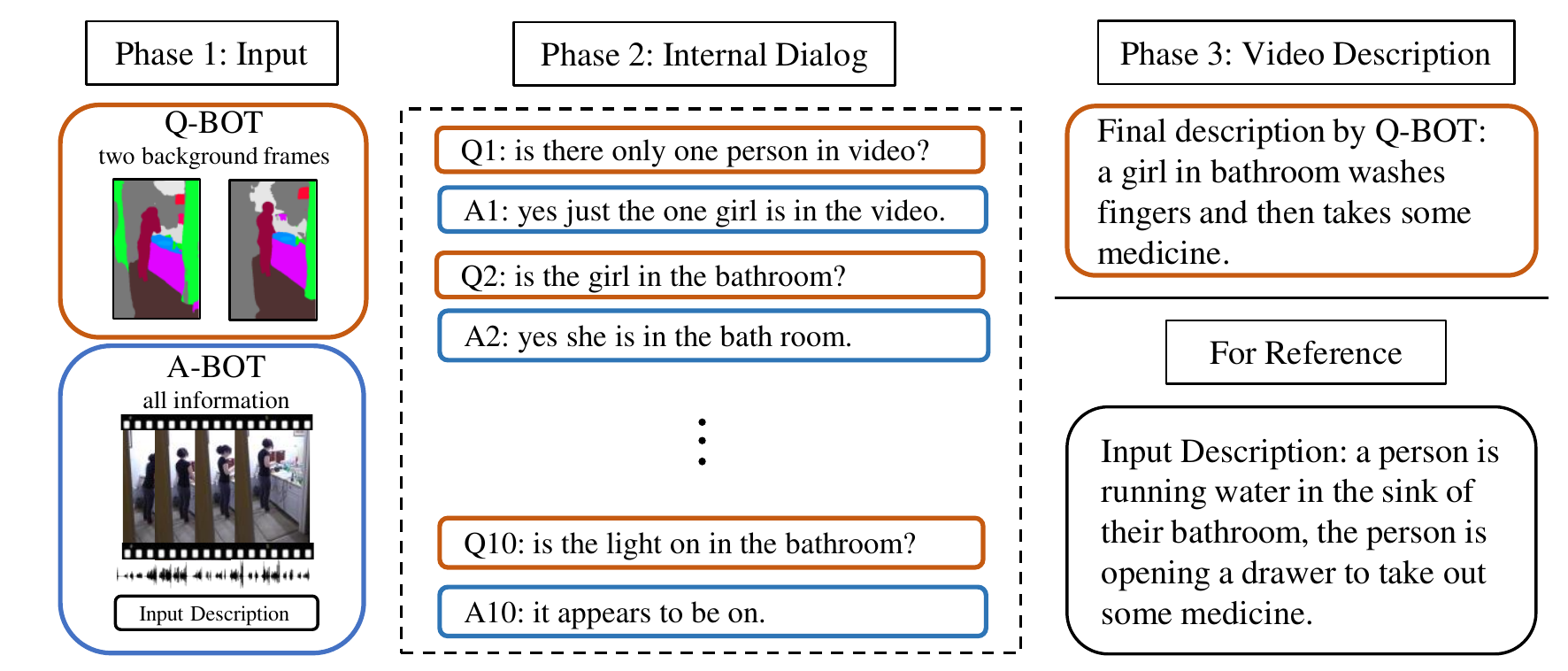}
    \caption{Unseen video description task via interpretable knowledge transfer between dialog agents. The task setup includes three phases, and the ultimate goal is for \emph{Q-BOT} to describe the unseen video mainly based on the dialog. The input description is also presented for reference. The difference between the input descriptions for \emph{A-BOT} and the final descriptions given by \emph{Q-BOT} reveals the actual knowledge gap due to the lack of direct access to the original video data.}
    \label{fig1:task}
\end{figure*}

Our task involves two agents, the questioning robot \emph{Q-BOT} and the answering robot \emph{A-BOT}. In practical scenarios such as smart homes, \emph{Q-BOT} could be the actual AI system, while \emph{A-BOT} plays the role of human users. Human users can naturally perceive all the information sources and answer questions related to the surrounding environment. In contrast, \emph{Q-BOT} (AI system) only has a sketchy perception of the general environment such that it will not see the entire home setting. 
The proposed task shares some similarities with the classic video captioning~\cite{pan2017video}, the visual dialog task~\cite{das2017visual}. Video captioning aims to generate a natural language summary of the video based on direct visual input, and visual dialog aims to answer a series of questions related to the visual content in the form of a dialog. 
Our task includes both components but differentiates them from multiple aspects. 
Firstly, the task inputs and motivations are different. While the previous tasks take the original complete visual data as input and seek to achieve better performance, our work intentionally provides the AI model with implicit visual input to exploit a trade-off between the performance and the visions.  
Secondly, the task goals are different. The original visual dialog task focuses on learning the AI systems to answer natural language questions. In contrast, our task emphasises to enable the AI models (\textit{i.e.}, Q-BOT) to achieve a concrete vision-related goal (\textit{i.e.}, video description) using the natural language dialog as a supplementary information source.

The concrete setup is illustrated in Figure~\ref{fig1:task}, which resembles the data collection process of the AVSD dataset~\cite{alamri2019audio}.
Initially, \emph{Q-BOT} takes as input two semantic segmented frames (\textit{i.e.}, semantic segmentation results of static video frames, thus no visible human faces) from the beginning and the end of the video.
\emph{A-BOT} has full access to the information of all modalities, including the entire video, audio stream, and the original video description sentences.
Afterwards, \emph{Q-BOT} has 10 chances to ask questions to collect necessary information of the video, and \emph{A-BOT} collaboratively provides answers to the questions. After 10 rounds of dialog, \emph{Q-BOT} is asked to summarize the unseen video based on the segmented visual input and the dialog history. Under our task setup, \emph{Q-BOT} learns to accomplish the video description task without directly seeing the video.

There are two principal considerations behind our task formulation that using the dialog as the supplementary information medium, instead of directly asking for the final descriptions from human users (\textit{i.e.}, \emph{A-BOT}). \textit{Firstly}, the overall descriptions directly given by humans are usually noisy and biased without given hints or templates in the sense that different humans may pay attentions to completely different parts given the same visual content~\cite{ungerleider2000mechanisms,kastner1999increased}. In contrast, the answers given by human users for specific questions are less biased. For a question like \textit{"How many persons are there in the video?"}, we can expect the answer to be a specific number in most cases.
\textit{Secondly}, from a higher-level perspective, AI systems have different objectives in practical scenarios, the question-guided dialog interactions help AI systems to better extract the necessary information required for accomplishing specific downstream tasks. For example, human users may want to create a better sound experience in their living rooms via the smart home system, which usually requires acoustic engineers to perform professional acoustic compensation based on the relative spatial relations among loudspeakers. For an acoustic expert, the process to acquire the spatial structure among loudspeakers is rather systematic via a succession of structured questions (\textit{e.g., How many loudspeakers do you have in the room? Are they placed in the corner close to the wall?}), while users may find it more challenging to directly provide the necessary spatial information. In this case, the AI system is expected to ask guided questions and to extract the necessary information from the answers provided. The above motivations inspire us to explore the possibilities of using the dialog as our primary choice for supplementing the insufficient visions.  

% Despite some similarities in the task setup, our task of unseen video description is different and more challenging compared to the traditional video captioning and visual dialog tasks. Specifically, while the previous tasks mainly emphasize the fulfillment of question answering, 
One of the unique challenges in our task is the effective knowledge transfer process from \emph{A-BOT} to \emph{Q-BOT}. 
% We choose to formulate the final objective in the form of natural language video descriptions as evaluations for the effectiveness of the internal knowledge transfer process.
To better illustrate and clarify the knowledge transfer process, we introduce the concepts of \emph{Input video descriptions for A-BOT} and \emph{Final output video descriptions by Q-BOT} (referred as \emph{Input Descriptions} and \emph{Final Descriptions} in the remaining of the paper). 
The main difference between the two types of video descriptions lies within the fact whether the person/agent has seen the entire video before giving the description. 
% \zy{In reference to} the original AVSD dataset~\cite{alamri2019audio}, the input video descriptions correspond to the video captions given by a human annotator (the role of A-BOT in our task) after watching the entire video. In contrast, the generated video descriptions correspond to the video summaries given by human annotators (the role of Q-BOT in our task) without directly seeing the video.
% , but only partial video frames and acquires the additional information from a conversation with another human annotator (the role of A-BOT in our task) who has already seen the entire video.
% The former description is the original video description given by human annotators after watching the \textit{entire} video, and is provided as part of the input information for \emph{A-BOT}. In contrast, the latter description is the output generated by \emph{Q-BOT} without seeing the original video. 
% We will refer the two types of descriptions as \emph{Input Descriptions} and \emph{Final Descriptions} in the remaining of the paper to avoid confusions.
% Figure~\ref{fig1:task} also includes concrete examples of the input descriptions and the final descriptions from the AVSD dataset~\cite{hori2019end,alamri2019audio}. 
The input descriptions example in Figure~\ref{fig1:task} contain more concrete details compared to the final descriptions. This fact demonstrates the knowledge and reasoning gap caused by the lack of direct access to the original video data, which also implies that although the natural language dialog could be an effective supplementary information source, it is rather challenging to completely alternate the incomplete visual information.
One significant step that leads to the successful accomplishment of our task is to reduce this gap by an effective knowledge transfer process between the agents.

\begin{figure*}[t]
    \centering
    \includegraphics[width=0.8\textwidth]{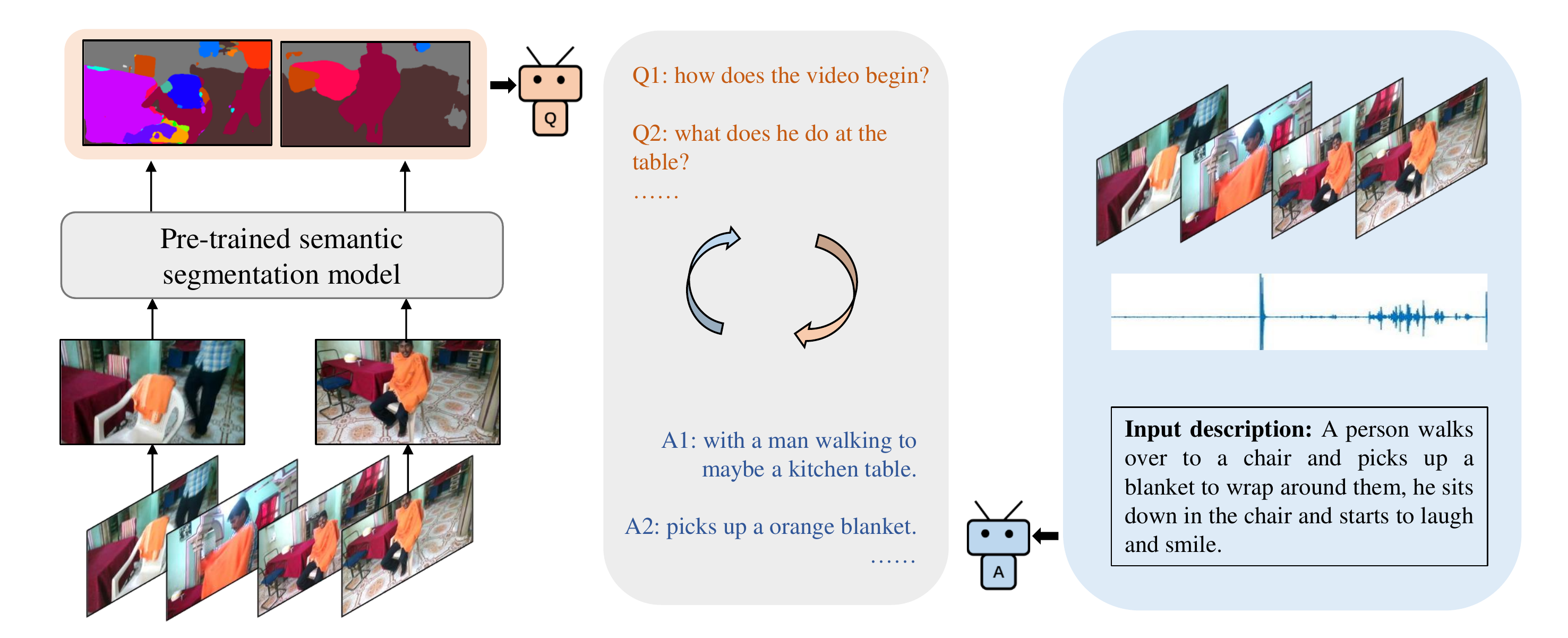}
    \caption{Different inputs for \emph{Q-BOT} and \emph{A-BOT} in shaded orange and blue boxes, respectively. For \emph{Q-BOT}, we extract the first and the last frames from the video clip and process the extracted frames using a pre-trained semantic segmentation model. We then input semantic segmented frames to \emph{Q-BOT}, which only provides a sketchy perception of the general environment and does not reveal any sensitive information. In contrast, \emph{A-BOT} has access to all the information, including entire videos, audio signals, and descriptions.}
    \label{fig:seg}
\end{figure*}

To accomplish the challenging task, we propose two different experimental settings with their corresponding unified QA-Cooperative networks. For the first experimental setting, \emph{Q-BOT} and \emph{A-BOT} generate questions and answers freely during the dialog interactions. We introduce a cooperative learning method with a dynamic dialog history update mechanism, which helps to transfer knowledge between the two agents effectively. Under this generative setting, we achieve promising performance and successfully transfer knowledge from \emph{A-BOT} to \emph{Q-BOT}. However, we also observe from the qualitative results that the generated internal dialog sometimes lacks clear logic and tends to be repetitive, which is a common issue in similar tasks~\cite{das2017learning}.
We believe a meaningful and informative internal dialog is in line with our final objective to obtain a precise final description. Therefore, to further enhance the internal reasoning abilities and the interpretability of the dialog agents, we propose an improved experimental setting where agents proceed internal dialog in a discriminative way, meaning that \emph{Q-BOT} and \emph{A-BOT} pick questions and answers from the given candidates. We then introduce an improved version of the QA-Cooperative network, and propose a different learning method with an internal selective mechanism to enhance the interpretability and quality of the internal dialog. 
With the improved setting and model design, we make significant improvements in the effectiveness of the knowledge transfer, leading to better final descriptions.
Through extensive experiments on the AVSD dataset~\cite{hori2019end,alamri2019audio}, we demonstrate: (a)~An effective knowledge transfer process between two agents via the proposed QA-Cooperative networks and learning methods. (b)~A meaningful and informative internal dialog indeed helps with our final objective to achieve better descriptions for unseen videos. (c)~Multiple data modalities and proposed model components contribute to the final performance.

This paper is \re{an extended work following~\cite{yzhu2020describing}}. Compared to the vanilla version~\cite{yzhu2020describing}, we incorporate a considerable amount of extension work from three aspects: the task setup, the methodology, and the experiments. \textit{For the task setup:} (a) we modify the initial task setup from~\cite{yzhu2020describing} to further enhance the security aspect of the task system. Specifically, the previous task setup of~\cite{yzhu2020describing} allows \emph{Q-BOT} to take two original RGB frames from the video as visual input. Although this task setup largely reduces the risk of exposing sensitive face images to AI systems, it can not ensure that the observed frames contain absolutely no biometric information. In this paper, we incorporate the semantic segmentation as pre-processing for \emph{Q-BOT} as shown in Figure~\ref{fig:seg}, and thus resolving the previous concern. (b) we add an improved discriminative setting for the internal dialog generation process, which brings us more interpretability for the internal reasoning process, as well as the improvement for the final descriptions. 
\textit{For the methodology:} (a) we propose an enhanced network architecture with modified question and answer decoders for the discriminative setting, which aims to enhance the quality and the interpretability of the generated dialog. (b) we incorporate an adapted internal selective mechanism from~\cite{qi2019two} for improved cooperative training that leads to better knowledge transfer and final performance.
\textit{For the experiments:} (a) we perform additional extensive quantitative and qualitative analysis for both the final descriptions and the dialog, which better interpret the internal process. 
(b) we conduct an simulated human test to evaluate the ability of \emph{A-BOT}.
(c) we achieve significant performance improvements for the final descriptions, raising the primary metric CIDEr~\cite{vedantam2015cider} from 22.9 to 27.1.

Our overall contributions for this work can be summarized as follows:
 \begin{itemize}
    %  \item We propose a novel and challenging task via two multi-modal dialog agents, which establishes a more secure and reliable task setup that allows AI systems to accomplish a concrete objective without direct access to sensitive personal information.
     \item \re{We propose a novel and challenging task that aims to describe an unseen video via two multi-modal dialog agents. The proposed task uses the natural language dialog as the supplementary information source for the incomplete visual input to address the potential privacy concerns. } 
    %  \item We introduce two settings for the internal dialog generation process that helps to effectively transfer knowledge between the two dialog agents.
    %  \item We propose two QA-Cooperative network framework designs for the corresponding settings, with a dynamic dialog history update mechanism and an internal reasoning mechanism to further enhance the interpretability of the models.
    \item \re{We introduce two QA-Cooperative framework designs for the internal dialog generation process. The proposed frameworks allow the two agents to fulfill the objective of unseen video description via a generative or discriminative internal dialog.}
     \item We conduct extensive experiments and analysis to show the effectiveness of the proposed methods for our novel task, achieving very competitive performance that beats multiple baselines. We also experimentally demonstrate the knowledge transfer process between two agents and the feasibility of using the natural language dialog as a supplementary for incomplete visual input.
 \end{itemize}

\section{Related Work}

\subsection{Image and Video Captioning}

Image and video captioning is a classic vision-language task that aims to textually describe the given image or video input. Previous work on image and video captioning usually provides the network models with direct access to original visual data~\cite{xu2015show,anderson2018bottom,chen2017sca,lu2017knowing,you2016image,wang2018reconstruction}.
% You \textit{et al.}~\cite{you2016image} tackle the image captioning task with attribute attention and Recurrent Neural Networks (RNNs), which combines top-down and bottom-up approaches by selectively attend to the semantic concept proposals.
Rennie \textit{et al.}~\cite{rennie2017self} formulate the image captioning problem with reinforcement learning \re{and optimize the problem using the self-critical sequence training}.
A disentangled framework is proposed by Wu \textit{et al.}~\cite{wu2018decoupled} to generalize image captioning models to describe unseen objects \re{for the zero-shot captioning task}. 
Transformer-based~\cite{Cornia_2020_CVPR} or attention-based~\cite{Pan_2020_CVPR,Guo_normalized} methods have also been adopted to tackle the problem.

Although the output of image and video captioning tasks is also the textual descriptions, our task has a different formulation with its focus on the internal knowledge transfer process between two agents via natural language dialog.

\subsection{VQA}

Visual Question Answering (VQA) is another popular vision-language task that aims to answer natural language questions relevant to the given visual data~\cite{antol2015vqa,lu2016hierarchical,xu2016ask,yang2016stacked,anderson2018bottom,das2017human,shih2016look,xiong2016dynamic,song2018explore,wu2020revisiting}.
% Dynamic memory networks are used for VQA in~\cite{xiong2016dynamic}.
% % Shih \textit{et al.}~\cite{shih2016look} map textual queries and visual features into a shared space for question answering.
% Lu \textit{et al.}~\cite{lu2016hierarchical} propose a hierarchical model with co-attention for VQA.  
More recent research works in VQA starts to bring the causality theory into the field.
Chen \textit{et al.}~\cite{chen2020counterfactual} propose a model-agnostic Counterfactual Samples Synthesizing (CSS) training scheme for robust question answering.
Agarwal \textit{et al.}~\cite{Agarwal_2020_CVPR} propose to reveal and reduce the spurious correlations for VQA models to achieve more robustness. Efforts are also made to achieve better performance and more diversity using techniques such as variational auto-encoders~\cite{wang2017diverse,chen2019variational}, attributes learning~\cite{yao2017boosting}, reinforcement learning~\cite{wu2018you,das2017learning} and pre-training~\cite{zhou2020unified}.

Most existing models for VQA aim to answer the given questions about the visual content as the task objective, while our work has a concrete objective (\textit{i.e.}, describing the unseen video) and uses the QA interactions as the medium for information transfer.

\subsection{Visual Dialog}

Unlike VQA that seeks to answer a single question, research works on visual dialog~\cite{das2017visual,de2017guesswhat,jain2018two,seo2017visual,wu2018you,gan2019multi,das2017human} extend the QA interactions into multiple rounds that form a complete and meaningful dialog with more internal logical relations. Several datasets have been collected~\cite{das2017visual,de2017guesswhat}. 
Most existing works in Visual Dialog emphasize the ability of AI to answer the questions, however, few researches have been done to exploit the other side of the problem, which the the question asking. Learning to ask meaningful and informative questions about the visual content is also an insightful research topic worth exploiting.
Jain \textit{et al.}~\cite{jain2018two} also look in to the problem of question asking.
% and propose a symmetric baseline for the visual dialog task from discriminative question and answer generation.
Qi \textit{et al.}~\cite{qi2019two} exploit the causality effect for the visual dialog task and propose two causal principles for improving existing models.  
Guo \textit{et al.}~\cite{Guo_2020_CVPR} introduce a Context-Aware Graph (CAG) neural network for the visual dialog task. 
Different attention mechanisms, such as the hierarchical attention~\cite{lu2016hierarchical}, question-guided spatial attention~\cite{xu2016ask}, stacked attention~\cite{yang2016stacked}, multi-step reasoning~\cite{song2018explore}, bottom-up and top-down attention~\cite{anderson2018bottom} have also been exploited and proven to be effective. 
Agarwal \textit{et al.}~\cite{agarwal2020history} recently study the role of history for visual dialog and reveal its potential shortcomings.
Works that leverage the advantages of pre-trained language models such as BERT~\cite{devlin2018bert} and then fine-tuned for visual dialog have also been exploied in~\cite{murahari2020large,wang2020vd}.

% Additionally, Das \textit{et al.}~\cite{das2017human} explore whether the existing attention mechanisms attend to the same regions as humans do, which brings the insights for the difference between real humans and machines.

% Most existing work above relies on direct features from images or videos to accomplish the task objective, while our task differentiates itself by acquiring supplementary information from an alternative information medium, \textit{i.e.}, the natural language dialog.
Despite some similarities in the task setup, our work takes the incomplete visual data as input and uses the dialog interactions to supplement the missing information. In addition, we shift the model focus from answering the questions to question asking.

\subsection{Audio Modality}

Audio modality is another important source of information that has gained research popularity in recent years. There have been emerging studies on combining audio and visual information for various downstream tasks such as the sound source separation~\cite{gao2018learning,owens2018learning,arandjelovic2018objects,zhao2018sound}, sound source localization~\cite{owens2018audio,gao2018learning,senocak2018learning} and audio-visual event localization~\cite{tian2018audio,wu2019dual,rouditchenko2019self,duan2020audio,wu2021explore,zhu2021learning}. 
Hu \textit{et al.}~\cite{hu2019deep} introduce Deep Multimodal Clustering for capturing the audio-visual correspondence.
Gao \textit{et al.}~\cite{Gao_2020_CVPR} propose to recognize actions in untrimmed video using audio as a preview mechanism to eliminate visual redundancies.

Audio-visual scene-aware dialog (AVSD)~\cite{alamri2019audio,hori2019end,schwartz2019simple} is another recently proposed task that resembles the visual dialog, it additionally incorporates audio signal compared to previous tasks and datasets~\cite{das2017visual,de2017guesswhat}. 
Hori \textit{et al.}~\cite{hori2019end} firstly propose an end-to-end model using multimodal attention-based video features to tackle the task. Alamri \textit{et al.}~\cite{alamri2019audio} further propose a benchmark for the AVSD task.
While the AVSD task still focuses on answering questions, our work seeks to describe the entire video, which requires the model to further extract useful information from the dialog.
In reference to the original AVSD dataset~\cite{alamri2019audio}, the input descriptions proposed in our work correspond to the video captions given by a human annotator (the role of A-BOT in our task) after watching the entire video. In contrast, the final descriptions correspond to the video summaries given by human annotators (the role of Q-BOT in our task) without directly seeing the video.

Our work uses the AVSD dataset for experiments and shows that the audio data is also an important information source that contributes to better performance for our task.
% \zy{In reference to the original AVSD dataset~\cite{alamri2019audio}, the input descriptions proposed in our work correspond to the video captions given by a human annotator (the role of A-BOT in our task) after watching the entire video. In contrast, the final descriptions correspond to the video summaries given by human annotators (the role of Q-BOT in our task) without directly seeing the video.}

\begin{table}[t]
\caption{Notations for the unseen video description task.}
\label{table1:notation}
\centering
\scalebox{1.0}{
\begin{tabular}{l}
\hline
$s$ - Final descriptions \\ \hline
$\mathcal{S}$ - Vocabulary  \\ \hline
$i (i \leq 10 )$ - Question-Answer round \\ \hline
$A$ - Audio data      \\ \hline
$V_A$ - Video data for A-BOT \\ \hline
$C$ - Input video descriptions    \\ \hline
$H_{i-1}$ - Existing dialog history at round $i$         \\ \hline
$p_i$ - $i$-th pair of question-answer\\ \hline
$q_i$ - $i$-th question \\ \hline
$a_i$ - $i$-th answer    \\ \hline 
$N_{\{q,a\}}$ - number of candidates \\ \hline
$V_s$ - start semantic segmented frame of the video \\ \hline
$V_e$ - end semantic segmented frame of the video \\ \hline
$x_{A,i}$ - input for \emph{A-BOT} at round $i$\\ \hline
$x_{Q,i}$ - input for \emph{Q-BOT} at round $i$  \\ \hline
$r_m$ - original data embedding for modal $m$ \\ \hline
$a_m$ - attended data embedding for modal $m$  \\ \hline
$d_m$ - dimension of the embedding for modal $m$ \\ \hline
$n_{\{C,H,q,a,s\}}$ - length of textual sequence\\ \hline
$m$ - modality notation, specified in context $m \in \{A,V,C,H,q,a\}$ \\ \hline
$e_{\{q,a\}}$ - embedding vector of candidates \\ \hline
$h_{\{v,av\}}$ - hidden states of LSTM\\ \hline
$c_{\{v,av\}}$ - cell states of LSTM\\ \hline
\end{tabular}}
\end{table}

\subsection{Cooperative Agents}
The research studies on dialog agents mainly have two categories. 
They either focus on maintaining a meaningful conversation~\cite{massiceti2018flipdial,wu2018you} or designing in a goal-driving manner to accomplish certain final objectives~\cite{das2017learning,lee2018answerer,strub2017end,zhang2018goal,rajendran2018learning,shukla2019should,guo2018dialog} (\textit{e.g.}, retrieve the image that the dialog is about from candidates).
Our work falls into the categories of the goal-oriented dialog systems.
Early works about goal-oriented dialog agents usually focus on the single modality of natural language and formulate the problem using machine techniques such as Markov chain process~\cite{singh1999reinforcement} and probabilistic learning~\cite{roy2000spoken}. 
More recent works seek to incorporate data of other modalities into the framework.
Das \textit{et al.}~\cite{das2017learning} train the dialog agents with reinforcement learning to select dialog-related images.
Guo \textit{et al.}~\cite{guo2018dialog} also propose to optimize the interactive dialog for retrieve images using deep reinforcement learning with a user simulator.  
An information theoretic algorithm for goal-oriented dialog is introduced in~\cite{lee2018answerer} to assist the question generation. 

The goal of our dialog system is to describe the unseen video. One of our challenges compared to the above works is the complexity of natural language descriptions, especially with incomplete visual input. Unlike the image retrieval task that aims to find the target image, the video descriptions are more various and difficult to evaluate. 
% \zy{Another challenge is the knowledge transfer process as we have mentioned in the introduction. Specifically, in reference to the original AVSD dataset~\cite{alamri2019audio}, the input descriptions proposed in our work correspond to the video captions given by a human annotator (the role of A-BOT in our task) after watching the entire video. In contrast, the final descriptions correspond to the video summaries given by human annotators (the role of Q-BOT in our task) without directly seeing the video. }

% \subsection{Saying the Unseen}

% \zy{Despite some similarities in the task setup, our task of unseen video description is different and more challenging compared to the traditional video captioning and visual dialog tasks. }

%%%%%%%%%%%%%%%%%%%%%%%%%%%%%%%%%%%%%%%%%%%%%%%%%%%%%

\section{Video Description via Dialog Agents}
We firstly present task formulations in Section~\ref{subsec:task_formulation}. The proposed QA-Cooperative networks for two settings are explained in Section~\ref{subsec:qac}. We then introduce their respective learning methods in Section~\ref{subsec:learning}.
Notations used in our formulations are listed in  Table~\ref{table1:notation}.

\subsection{Task Formulation}
\label{subsec:task_formulation}
\subsubsection{Video Description}
For the proposed unseen video description task, our primary goal is for \emph{Q-BOT} to describe an unseen video with a sentence $s = (s_1, s_2, ..., s_{n_s})$ in $n_s$ words after 10 rounds of QA interactions.
Each word $s_k$ arises from a vocabulary $\mathcal{S}$. At $i$-th round of QA interaction, \emph{A-BOT} takes the video data, audio signals, input description and the existing dialog history as input.
Denote the input data to be $x_{A,i} = (A, V_A, C, H_{i-1})$, where 
$A$ is the audio data, $V_A$ is the complete video data, $C$ is the input video descriptions, and the dialog history $H_{i-1} = \{p_1, ..., p_{i-1}\}$ with $p_{i}$ to be the QA pairs $p_{i} = (q_{i}, a_{i})$. For \emph{Q-BOT} at the same round $i$, we extract the first and last frames from the video, and then perform semantic segmentation on these two frames using pre-trained models to obtain $V_s$ and $V_e$. The semantic segmented frames $V_s$ and $V_e$ eliminate the possibilities to reveal recognizable human faces to \emph{Q-BOT}, as shown in Figure~\ref{fig:seg}. The input data for \emph{Q-BOT} is $x_{Q,i} = (V_s, V_e, H_{i-1})$.
The final description task for \emph{Q-BOT} is formulated as the inference in a recurrent model with the joint probability given by:
\begin{equation}\label{eq:joint_probability}
    p(s|x_Q) = \prod_{k=1}^{n_s}p(s_k|s_{<k},x_Q),
\end{equation}
\noindent where we maximize the product of conditionals for each word in description $s$, given the input at 10-th round $x_Q$.
From Eqn. (1), the core is how to generate a better dialog history $H$ in $x_Q$. Next, we illustrate how to generate the internal dialog in two ways.

\subsubsection{Generative Dialog}

One intuitive and straightforward way to formulate the internal dialog process is for both agents to directly generate the questions and answers. In this case, \emph{Q-BOT} and \emph{A-BOT} have the flexibility to freely ask questions and to provide answers. The generated questions and answers are formulated in a similar way as the final description. At $i$-th round of QA interactions, the $i$-th question $q_i$ is given by:
\begin{equation}
    p(q_i|x_{Q,i}) = \prod_{k=1}^{n_q}p(q_{i,k}|q_{i,<k},x_{Q,i}),
\end{equation}
\noindent where $n_q$ is the number of words for the $i$-th question.
Similarly for \emph{A-BOT}, the $i$-th answer is generated following the same equation by replacing $Q$ and $q$ with $A$ and $a$, respectively.
% \begin{equation}
%     p(a_i|x_{A,i}) = \prod_{k=1}^{n_a}p(a_{i,k}|a_{i,<k},x_{A,i}),
% \end{equation}
Under this setting, the information is cooperatively exchanged through the dialog $H$. However, it is more challenging to guarantee the quality of the generated dialog due to the lack of supervision for the generation process.

\subsubsection{Discriminative Dialog}

Another way to obtain the internal dialog is to provide possible candidates for \emph{Q-BOT} and \emph{A-BOT} to choose from. More specifically, $q_i$ and $a_i$ are picked from potential candidates $\{q_i^1, q_i^2,..., q_i^{N_q}\}$ and $\{a_i^1, a_i^2,..., a_i^{N_a}\}$ by \emph{Q-BOT} and \emph{A-BOT}, respectively. 
Those candidates are selected from the dataset.
During inference, all the questions and answers from testing dialog are provided as candidates for \emph{Q-BOT} and \emph{A-BOT}. 
During training, we provide 100 questions and 100 answer candidates for each case. All the ground truth questions/answers, except those provided as input, are included in 100 candidates.
Other candidates are randomly selected from the training set. Additionally, all the candidates are provided in pairs.
In other words, for each question from the question candidates for \emph{Q-BOT}, we also include its corresponding ground truth answer as an answer candidate for \emph{A-BOT}. 
However, it should be mentioned that if a new question other than the ground truth ones is picked by \emph{Q-BOT}, the picked question may not be valid (\textit{i.e.}, the question may be irrelevant to the given video). In this case, there may not exist valid answers.

Considering the fact that comparing all the candidates at each QA round is very time-consuming during inference, we deploy a two-phase selection mechanism: cluster selection and question/answer selection. 
\emph{Q-BOT} and \emph{A-BOT} firstly select the pre-clustered question or answer type, and then pick the specific candidates from the previously chosen cluster.
For the question and answer types, we represent each question and answer from the testing set with the Glove embedding~\cite{pennington2014glove} and use the K-Means algorithm to cluster sentences into 10 classes. 
Specifically, we use the pre-trained Glove model to convert the each word from sentences into feature vectors and perform the clustering on the obtained sentence embeddings.
During inference, the agent first picks a sentence cluster, and then further choose a concrete sentence within the cluster.

A discriminative setting for the internal dialog helps with alleviating the bias commonly existing in vision-language models and provides more interpretability for the internal dialog process.
Overall, we experimentally demonstrate that both internal dialog settings are viable for our unseen video description task in Section~\ref{sec:exp}. The generative setting may be more flexible for general deployment, while the discriminative setting leads to the stronger final performance and better reveals the internal reasoning process. 

\subsection{QA-Cooperative Networks}
\label{subsec:qac}

\subsubsection{Model Components}

Our QA-Cooperative networks include multiple model components, which are presented in detail in this section. We focus on the situation at $i$-th round of dialog.

\noindent \textbf{Components of Q-BOT.}
The \emph{Q-BOT} contains a visual LSTM~\cite{hochreiter1997long} that processes the input frames, a history encoder that gathers information from the dialog history, a question decoder for generating questions, and a description generator that finally describes the unseen video.

\noindent \textit{Visual LSTM.} It is an LSTM with 2 units, this component takes the attended visual embedding $a_{V,s} \in \mathbb{R}^{d_V}$ and $a_{V,e}\in \mathbb{R}^{d_V}$ as input, the hidden states and cell states $(h_v, c_v)$ from this LSTM is used as the initial states for the question decoder and the final description generator. \emph{Q-BOT} uses this component to process the visual information from the two semantic segmented frames.

\noindent \textit{History Encoder.} It consists of a linear layer and a single layer LSTM. We start with a list of one-hot word representations for a QA pair. 
The longest QA pair of length is selected, the other pairs are zero-padded to fit the maximum length. The LSTM is used to obtain the pair-level embedding $r_{H,i-1} \in \mathbb{R}^{n_T \times n_H \times d_H}$, where $n_T$ denotes the number of QA pairs in the current dialog history (\textit{i.e., $i-1$}).

\noindent \textit{Question Decoder.} 
The question decoder has different functions for generative and discriminative dialog settings.
For the generative dialog setting, the question decoder is formed by an LSTM. It takes the attended history embedding $a_{Q,H,i-1} \in \mathbb{R}^{d_H} $ as input, with initial states $(h_0, c_0) = (h_{v}, c_{v})$. 
The question generator generates the new question $q_i$ \re{based on the} $i$-th question in the ground truth dialog for the generative dialog setting.
For the discriminative dialog setting, the question decoder is used to calculate the score based on the similarity between the question candidate embedding vector and the input for the question decoder $Score_{i}^n = <e_{q_i^{n}}, a_{Q,H,i-1}>$, where the notation $<.>$ denotes the inner product. The score computing is applied to each possible candidate. The scores are then processed by a Softmax operation to obtain the probabilities of all the candidates. The candidate with the highest probability is selected as the final $q_i$.

\noindent \textit{Description Generator.} This LSTM generator generates the final description $s$ for the unseen video based on 10 rounds of QA interaction history and the two semantic segmented frames given in the first phase. When $i=10$, the generator computes the following conditional probabilities based on the input, which is the attended history embedding $a_{A,H,10} \in \mathbb{R}^{d_H}$ including 10 rounds of QA interactions:
\begin{equation}
    % (h_0, c_0) = (h_v, c_v)
    p(s_k|s_{k-1}, h_{k-1}, x_Q) = g(s_k, s_{k-1}, h_{k-1}, x_Q),
\end{equation}
\noindent 
where $h_{k-1}$ is the hidden states from the previous $k-1$ step. 
The LSTM $g$ predicts the probability distribution $p(s_k|s_{k-1}, h_{k-1}, x_Q)$ over words $s_k \in \mathcal{S}_k$, conditioned on the previous words $s_{k-1}$. The final probability distribution for the description is obtained by transforming the output of the LSTM by a FC-layer and a Softmax operation.

\noindent \textbf{Components of A-BOT.}
The \emph{A-BOT} contains an audio-visual LSTM that processes the audio and visual input, the same history encoder as \emph{Q-BOT} that gathers information from the dialog history, an input description encoder that processes the input video descriptions, and an answer decoder used to generate answers.

\noindent \textit{Audio-visual LSTM.} It is an LSTM with $d+1$ steps, where $d$ is the number of visual frames visible to \emph{A-BOT}. The extra one step is for processing the audio input.  It takes the attended audio embedding $a_A \in \mathbb{R}^{d_A}$ and the attended visual embedding $a_{V,j} \in \mathbb{R}^{d_V}$ with $j = \{1, ...,d\}$ as input. The hidden states and cell states $(h_{av}, c_{av})$ generated from this LSTM are used as the initial states input to the answer decoder.. 
This component is for \emph{A-BOT} to process the audio and visual information in addition to the cross-modal attention.

\noindent \textit{Input Description Encoder.} It consists of the same structure as the history encoder with a linear layer and an LSTM. The input description embedding $r_C \in \mathbb{R}^{n_C \times d_C}$ is obtained from the last hidden state of the LSTM. This component is designed for \emph{A-BOT} to encode the input descriptions.

\noindent \textit{History Encoder.} It is the same encoder as the one for \emph{Q-BOT} since the history is a common input visible for both agents.

\noindent \textit{Answer Decoder.} Similar to the question decoder for \emph{Q-BOT}, this component is used to get the answer $a_i$ for question $q_i$.
The only difference is that this answer decoder takes the concatenation of the attended history embedding $a_{A,H,i-1} \in \mathbb{R}^{d_H}$, the attended input description embedding $a_c \in \mathbb{R}^{d_C}$ and the newly generated question embedding $r_{q,i}$ as input, with initial state $(h_0, c_0) = (h_{av}, c_{av})$. 
The output is the answer $a_i$ for the given question. 
The newly generated QA pair at $i$-th round is obtained by combining the $i$-th question and answer.

\begin{figure*}[t]
    \centering
    \includegraphics[width=0.97\textwidth]{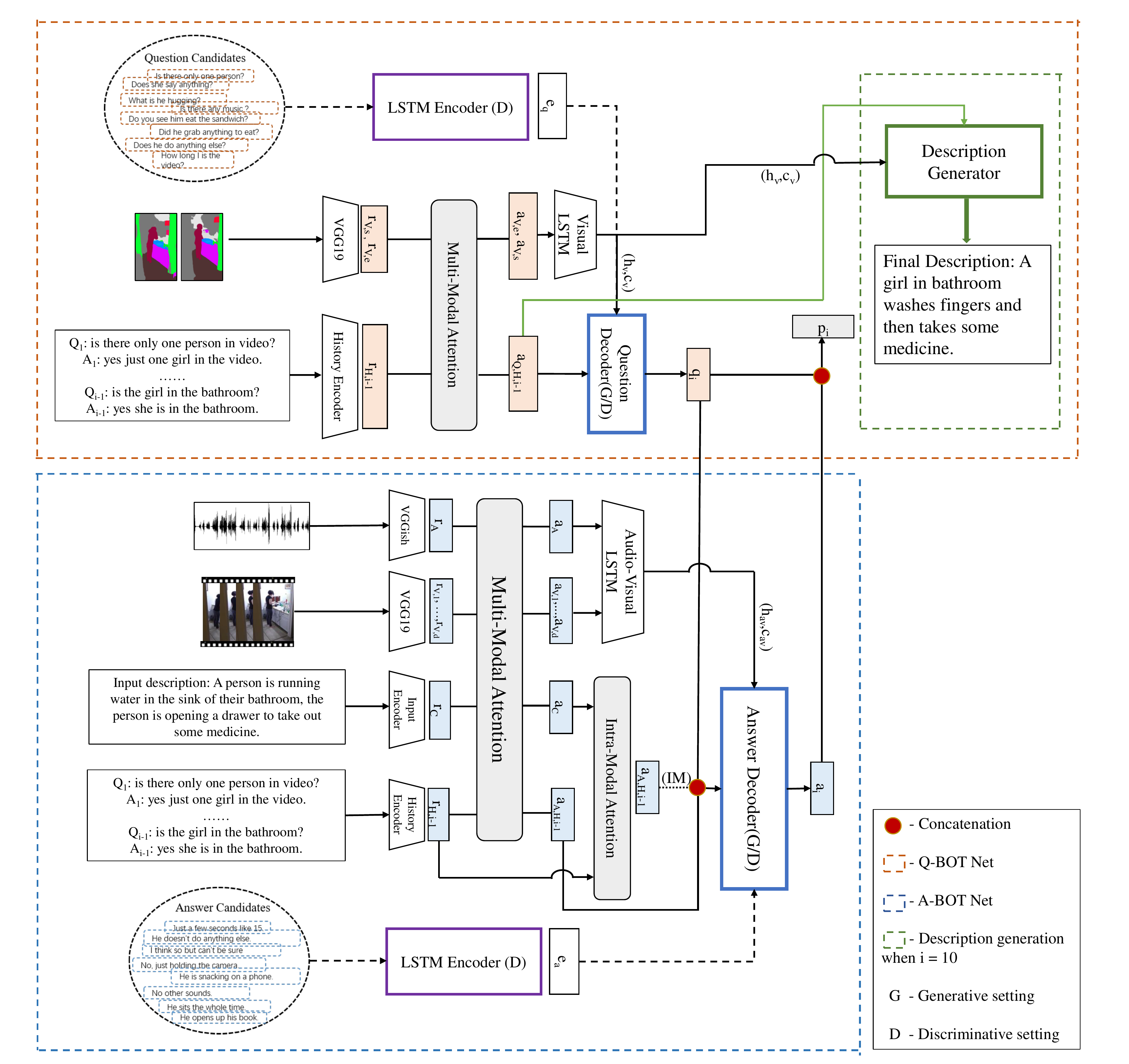}
    \caption{QA-Cooperative network at QA round $i$. History dialog $H$ is a common input for both agents. The model components for \emph{Q-BOT} are in orange color boxes, while those for \emph{A-BOT} are in blue. 
    % When the dialog history contains 10 rounds of QA pairs, \emph{Q-BOT} generates the final descriptions via the description generator.
    The dashed lines represent the processing of the question and answer candidates uniquely for the discriminative setting, while the solid lines are operations for both generative and discriminative settings.}
    \label{fig2:qac}
\end{figure*}

\noindent \textbf{Attention modules} 
Since the dialog is a key information source in our task to supplement the missing visual input, we propose two different attention mechanisms for processing the information contained in the dialog history: (1) the multi-modal (MM) attention among visual, audio, and textual modalities, and (2) the intra-modal (IM) attention between dialog history and another textual sequence. 

\noindent \textit{MM Attention.} We use the factor graph attention mechanism proposed in~\cite{schwartz2019factor} for MM attention module. For \emph{A-BOT}, this MM attention module takes the audio embedding $r_A$, visual embedding $r_{V,j}$ with $j = \{1,...,d\}$, input description embedding $r_C$ and the history embedding $r_{H,i-1}$ as input. Each visual frame is treated as an individual modality as in~\cite{schwartz2019factor}. The output of this multi-modal attention module are the attended audio embedding $a_A$, the attended visual embedding $a_{V,j}$ with $j = \{1,...,d\}$, and the attended history embedding $a_{Q,H,i-1}$. Similarly for \emph{Q-BOT}, we have the attended visual embedding $a_{V,s}$, $a_{V,e}$ and the attended history embedding $a_{A,H,i-1}$ as output, after taking their original embedding $r_{V,s}$, $r_{V,e}$ and $r_{H, i-1}$ as input. Note that the history embedding $r_{H,i-1}$ before the MM attention module is the same for \emph{Q-BOT} and \emph{A-BOT} because of the shared history encoder, but the attended history embedding becomes different due to different inputs for two agents.

\noindent \textit{IM Attention.} We adopt a softmax attention consisting of the concatenation and dot product operations between the dialog history embedding $r_{H,i-1}$ and the question embedding $r_{q,i}$ as the IM attention.

\subsubsection{QA-Cooperative Framework}
 
The architecture for our proposed QA-Cooperative networks is presented in Figure~\ref{fig2:qac}. The main difference in network architecture for two internal dialog settings is the design of question/answer encoder/decoder as explained in model components.

In general, the dialog history consisting $i-1$ QA pairs is a common input for both agents since it is visible to both agents in practical scenarios. \emph{Q-BOT} processes the visual input (two semantic segmented frames) and the dialog history input via VGG19 and the history encoder to obtain the visual and history embedding $r_{V,s}$, $r_{V,e}$, and $r_{H,i-1}$ respectively. 
They are later processed by the MM attention module to obtain the attended embedding $a_{V,s}$, $a_{V,e}$ and $a_{H,i-1}$. 
The attended visual embedding
$a_{V,s}$ and $a_{V,e}$ are then fed into the Visual LSTM to get the states output $(h_v, c_v)$. The question LSTM decoder takes the attended history embedding $a_{H,i-1}$ as input and outputs the $i$-th question $q_i$. Similarly for \emph{A-BOT}, it takes the video frames, audio signal and original input descriptions as input. These modalities of input data are processed by VGG19, VGGish~\cite{hershey2017cnn} and the input description encoder to obtain their respective embedding $(r_{V,1},..., r_{V,d}$, $r_A$ and $r_C$. The MM attention module takes those embedding and the history embedding $r_{H,i-1}$ as input and outputs the attended embedding vectors. While the attended audio and visual embedding vectors are processed by the audio-visual LSTM to obtain the states $(h_{av}, c_{av})$, the attended history embedding $a_{H,i-1}$ is fused with $i$-th question embedding $r_{qi}$ to form the input for answer decoder. After having obtained $i$-th answer $a_i$, the $q_i$ and $a_i$ are used to form the $i$-th QA pair and to update the existing dialog history. When the dialog history includes 10 QA pairs, \emph{Q-BOT} generate the final descriptions.

\subsection{Cooperative Learning}
\label{subsec:learning}

We propose to learn the proposed QA-Cooperative networks with corresponding cooperative learning methods, which have different emphasis for the generative and discriminative dialog settings.

\subsubsection{Dynamic History Update Mechanism}

The dialog history is \re{an important supplementary} information source for \emph{Q-BOT} to describe the unseen video in our task. 
Under the generative internal dialog setting, we therefore propose to update the dialog history in a dynamic way~\cite{yzhu2020describing}. To be more concrete, we maintain the embedding dimension of the newly generated QA pair equal to the dimension of the existing dialog history to emphasize the new information at each QA round. We deploy a linear layer to reduce the dimension of the existing dialog history $d_{H_{i-1}}$ to the size of the current QA pair $d_{p_i}$, and then concatenate the dialog history embedding with the embedding obtained for the $i$-th QA round.

\subsubsection{Internal Selective Mechanism}
\label{subsec:reasoning}

We introduce a mechanism to \re{improve the quality of the internal dialog under the discriminative setting where the dialog agents are expected to select appropriate questions and answers from given candidates}. 

It is usually more difficult to guarantee the quality of internal dialog under the generative internal dialog setting, since no internal evaluations are implemented on the dialog round level. However, we believe that the quality of the internal dialog is a significant factor that contributes to a better final description, which is also experimentally demonstrated in our experiments in Section~\ref{sec:exp}. In order to improve the quality of the dialog for the discriminative setting, we propose an internal selective mechanism via the sparse annotations similar to~\cite{qi2019two}. 
% Specifically, we compute the internal loss using the sparse annotations during the internal selection process as $\mathcal{L}_{internal} = \sum_i y_i, y_i \in \{0,1\}$. $y_i$ is 1 if the selected question/answer is not the ground truth one at each round, 0 otherwise. 
Specifically, it can be considered as a pre-training stage during which the agents learn to reason. We compute the internal loss using the sparse annotations during the internal selection process as:

$\mathcal{L}_{internal} = \sum_i y_i \: log \: softmax(logit_i), y_i \in \{0,1\}$. 

\noindent $y_i$ is 1 if the selected question/answer is not the ground truth one at each round, 0 otherwise. The spare annotation refers to the fact the we only consider the binary selections of the ground truth questions and answers while computing the internal loss, which contrasts to the idea of considering their dense relevance scores~\cite{qi2019two}.

Overall, we have two loss terms during the entire training process, \textit{i.e.}, $\mathcal{L}_{internal}$ and $\mathcal{L}_{CE}$. $\mathcal{L}_{internal}$ is the loss computed using sparse annotations in order to enhance the reasoning ability of the agents. $\mathcal{L}_{CE}$ is the cross-entropy loss on the probabilities of the final description. Thus we combine the internal loss term to enhance the reasoning ability of two agents and improve the quality of the internal dialog: 
$\mathcal{L} = \lambda \mathcal{L}_{internal} + (1-\lambda) \mathcal{L}_{CE}$.
$\lambda$ is the weight for the internal loss term. We empirically set $\lambda$ to be 0.1 in our experiments.
In our training process, we first optimize our model using the above loss fuction. 
To stabilize the optimization, we train our model using only the cross-entropy loss $\mathcal{L}_{CE}$ in the last several training epochs.
% , when we observe no further performance improvements using Eqn.~(\ref{eq:optimization}).

%%%%%%%%%%%%%%%%%%%%%%%%%%%%%%%%%%%%%%%%%%%%%%%%%%%%%%%%%%%%%%%%%%
\section{Experiments}
\label{sec:exp}

\subsection{Dataset}

We evaluate the proposed method on the AVSD dataset~\cite{hori2019end,alamri2019audio}. The data collection process reassembles our task setup where two Amazon Mechanical Turks (AMT) play the role of Questioner and Answerer. The Questioner was shown only the first, middle, and last static frames of the video, while the Answerer had already watched the entire video, including the audio stream and the original input descriptions. After having a conversation about the events that happened between the frames through 10 rounds of QA interactions, the Questioner is asked to summarize the entire video. 
The AVSD V0.1 is split into 7659 training, 734 prototype validation and 733 prototype testing dialog, each dialog consists of 10 rounds of question and answer pairs, accompanying the corresponding video clip, audio signals and input descriptions. Our experiments are performed on the provided training, validation and testing split.

\subsection{Implementation}

\subsubsection{Evaluation Metrics.}
% \noindent \textbf{Evaluation Metrics.}
The BLEU1-4~\cite{papineni2002bleu}, METEOR~\cite{banerjee2005meteor}, SPICE~\cite{anderson2016spice}, ROUGE\_L~\cite{lin2004rouge}, and CIDEr~\cite{vedantam2015cider} are used as the quantitative evaluation metrics for our generated final descriptions. CIDEr, which measures the similarity of a sentence to the consensus, is our primary metric for evaluations.
For the internal dialog interactions between two agents under the discriminative setting, we further compute and analyze the ground truth question and answer selection ratios during training as additional quantitative evaluations.

\subsubsection{Data Representations}

Our cooperative dialog agents have multiple modalities of data input including visual, audio, and textual data. For the visual data of \emph{A-BOT}, we take the video representations extracted from the last conv layer of a VGG19 as input. We sample four equally spaced frames from the beginning of the original video, and each frame representation is of dimension $49 \times 512$, where spatial and visual embedding dimensions are 49 and 512, respectively. For the visual input of \emph{Q-BOT}, which only serves the purpose of sketchy perception of the general environment, we begin with getting the first and last frames of the video, and then perform the semantic segmentation using the pre-trained PSPNet with ResNet-50 on the ADE20K dataset~\cite{zhou2017scene,zhou2018semantic}. The segmentation result images are used to extract the representations following the same procedure as for \emph{A-BOT}, the final representation is of dimension $28 \times 512$.
For the audio modal, we obtain the 256-dim audio feature via VGGish~\cite{hershey2017cnn}. For the textual representations, we extract the language embedding from the last hidden state of their corresponding LSTM. The dimensions are $d_C = 256$, $d_q = 128$, $d_a = 128$ and $d_H = 256$.

\subsubsection{Test Settings}
\label{subsec:test_setting}
During our test, the performance of \emph{Q-BOT} is evaluated at each QA round-level. In other words, each dialog is split into ten independent evaluation cases with the starting round number ranges from 1 to 10. For example, if the start round number $i $is 1, then no existing dialog history is given to \emph{Q-BOT} and \emph{A-BOT}, they will generate all the 10 questions and answers by themselves. However, if the start round number $i$ equals $6$, then five rounds of QA pairs are given to two agents as the existing history, in which case, \emph{Q-BOT} still has another five changes to freely ask questions. For a given video, testing with different start round numbers is independent, resulting in 10 different test cases. Therefore, for the 733 videos from the test set of the AVSD dataset~\cite{hori2019end}, we have in total 7330 different test cases. We refer to this testing process as the standard test setting. We also conduct a "strong baseline" experiment with the full ground truth dialog provided as input. For the strong baseline situations, there are only 733 test cases due to the fact that the entire dialog history is provided. 

\begin{table*}[t]
\begin{center}
\caption{Quantitative experimental results of the unseen video description task. \textit{HIS Att} stands for \textit{History attention}. \emph{G} and \emph{D} denote the Generative or Discriminative dialog settings.
The experiments are split into multiple groups, the group for \emph{A-BOT} helps to understand the knowledge gap between two dialog agents. 
We obverse that both \emph{A-BOT} and \emph{Q-BOT} from the proposed QA-Cooperative networks successfully transfer the knowledge by achieving very competitive performance that beats multiple baselines.
For better visualization, we mark the scores in descending order with the \textcolor{newgreen}{green}-\textcolor{newblue}{blue}-black color gradient.}
\scalebox{0.97}{
\label{tab:results}
\begin{tabular}{|c|c|c|c|c|c|c|c|c|c|c|}
\hline
Group & Method & HIS Att  & BLEU1  & BLEU2    & BLEU3   & BLEU4        & METEOR  & SPICE       & ROUGE\_L      & CIDEr    \\ \hline
\multirow{6}{*}{\begin{tabular}[c]{@{}c@{}}A-BOT\end{tabular}} & Hori \textit{et al.}~\cite{hori2019end}   & - & \textcolor{newblue}{34.2}    & 17.1    & 8.4     & 4.8    & 11.5 & 11.4 & 24.9   & 20.7    \\ \cline{2-11} 
 & S. \textit{et al.}~\cite{schwartz2019simple}   & -    & 32.1    & 16.2     & 8.7      & 5.1    & 12.1  &  11.6  & 27.6    & 21.6   \\ \cline{2-11} 
  & S. \textit{et al.}~\cite{schwartz2019simple}   & IM    & 33.8    & 16.9     & 9.1      & 5.3    & \textcolor{newblue}{12.7} & 11.8  & 27.7    & 22.7   \\ \cline{2-11} 
 & S. \textit{et al.}~\cite{schwartz2019simple} & MM      & 33.8    & \textcolor{newblue}{17.6}    & \textcolor{newblue}{9.9}     & \textcolor{newblue}{5.9}      & \textcolor{newblue}{12.9} &  \textcolor{newblue}{13.5} & \textcolor{newblue}{28.5}      & \textcolor{newblue}{25.6}   \\ \cline{2-11} 
  & Ours    & IM   & \textcolor{newgreen}{\textbf{37.9}} & \textcolor{newgreen}{ \textbf{21.6}} & \textcolor{newblue}{ 12.5}   & \textcolor{newblue}{7.6}    & \textcolor{newgreen}{\textbf{15.2}} & \textcolor{newgreen}{\textbf{18.5}} & \textcolor{newblue}{31.1}          & \textcolor{newblue}{38.1}          \\ \cline{2-11} 
 & Ours    & MM      & \textcolor{newblue}{ 37.5}          & \textcolor{newblue}{ 21.5}          & \textcolor{newgreen}{\textbf{12.9}} & \textcolor{newgreen}{\textbf{8.2}} & \textcolor{newgreen}{\textbf{15.2}} & \textcolor{newblue}{17.9} & \textcolor{newgreen}{\textbf{31.2}} & \textcolor{newgreen}{\textbf{39.3}} \\ \hline \hline
 
\multirow{3}{*}{\begin{tabular}[c]{@{}c@{}}Q-BOT\\ Basic baselines\end{tabular}}  
& Ours w/o dialog    & -    & 28.1     & 12.4    & 6.5     & 3.5     & 11.0 &  8.2   & 25.0    & 14.2      \\ \cline{2-11}
& Ours    & IM    & 31.8     & 15.6    & 8.1     & 4.5      & 11.6 &  11.0   & 25.8    & 18.0      \\ \cline{2-11} 
 & Ours    & MM     & 33.1    & 16.0     & 8.3      & 5.1     & 12.5 & 11.2  & \textcolor{newblue}{27.8}      & 22.1     \\ \hline
\multirow{2}{*}{\begin{tabular}[c]{@{}c@{}}Q-BOT\\ Strong baselines \end{tabular}}                    
& \begin{tabular}[c]{@{}c@{}}Ours\\(full GT HIS)\end{tabular}   & IM  & \textcolor{newblue}{33.5}   & \textcolor{newblue}{17.0}     & 8.9           & 5.4     & 12.7 &  11.5 & 27.0    & 21.2      \\ \cline{2-11} 
  & \begin{tabular}[c]{@{}c@{}}Ours\\(full GT HIS)\end{tabular}   & MM      & \textcolor{newgreen}{\textbf{34.7}}          & \textcolor{newgreen}{\textbf{18.4}}          & \textcolor{newblue}{10.2}           & \textcolor{newblue}{6.1}          &  \textcolor{newgreen}{\textbf{13.6}}  &      \textcolor{newgreen}{\textbf{14.2}} &      \textcolor{newgreen}{\textbf{28.7}}    &     \textcolor{newblue}{ 25.9 } \\ \hline
  
\multirow{8}{*}{\begin{tabular}[c]{@{}c@{}}Q-BOT\\ Cooperative\end{tabular}}  
&\begin{tabular}[c]{@{}c@{}}Ours-G\\ (pre-trained)\end{tabular} & MM     & 31.4 & \textcolor{newblue}{17.1}          & 9.2         & 5.4         & 12.7 & 11.4 & 27.1         & 21.3         \\ \cline{2-11} 
  & Our QA-C~\cite{yzhu2020describing}  & IM       & \textcolor{newblue}{33.3} & \textcolor{newblue}{17.0} & 9.1 & 5.4 & 12.6& 11.7 & 27.3 & 21.3 \\  \cline{2-11} 
    & Our QA-C~\cite{yzhu2020describing}  & MM      & \textcolor{newblue}{33.3} & \textcolor{newblue}{17.3} & \textcolor{newblue}{9.5} & \textcolor{newblue}{5.5} & \textcolor{newblue}{12.8} & \textcolor{newblue}{12.4} & \textcolor{newblue}{27.9} & \textcolor{newblue}{23.1} \\  \cline{2-11} 
  & Our QA-C(G)  & IM      & 31.8 & 16.2 & 9.1 & 5.3 & 12.7& 11.6 & 27.0 & 21.2 \\  \cline{2-11} 
&Our QA-C(G) & MM      &  32.4 &   16.3   & \textcolor{newblue}{9.7}     & 5.4       & \textcolor{newblue}{12.9} & 11.3 & 27.7         & \textcolor{newblue}{22.9}   \\  \cline{2-11}
  & Our QA-C(D)  & IM      & \textcolor{newblue}{33.8} & \textcolor{newblue}{17.7} & \textcolor{newblue}{9.7} & \textcolor{newblue}{5.9} & \textcolor{newblue}{12.8} & \textcolor{newblue}{13.2} & \textcolor{newblue}{28.2} & \textcolor{newblue}{26.1} \\  \cline{2-11}
    & Our QA-C(D)  & MM    & \textcolor{newgreen}{\textbf{34.7}} & \textcolor{newblue}{18.0} & \textcolor{newblue}{10.2} & \textcolor{newblue}{6.1} & \textcolor{newblue}{13.2} & \textcolor{newblue}{13.6}& \textcolor{newblue}{28.6} & \textcolor{newblue}{27.1} \\  \cline{2-11}
    & \re{QA-C(D) w/ simulated A}  & MM    & \textcolor{newblue}{34.3} & \textcolor{newgreen}{\textbf{18.4}} & \textcolor{newgreen}{\textbf{10.3}}  & \textcolor{newgreen}{\textbf{6.3}} & \textcolor{newblue}{13.4} & \textcolor{newblue}{14.1} & \textcolor{newblue}{28.6} & \textcolor{newgreen}{\textbf{27.6}} \\    
\hline

\end{tabular}}
\end{center}
% \label{tab:results}
\end{table*}

% \noindent \textbf{Training Details.}
\subsubsection{Implementation Details}

The description generator of our proposed QA-Cooperative networks is trained using a cross-entropy loss on the probabilities $p(s_k|s_{<k}, x_Q)$ on the final descriptions. All the components are jointly trained in an end-to-end manner. The total amount of trainable parameters is approximately 19M for the generative dialog setting and 12M for the discriminative dialog setting. We use the Adam optimizer with a learning rate of 0.001 and a batch size of 64 for training. During training, we evaluate the performance on the validation set with a perplexity metric. The training stops after two consecutive epochs with no improvement in the perplexity metric. 
% The training time is approximately 5hr and 4hrs with a GeForce GTX 1080Ti GPU for generative and discriminative settings, respectively.

\subsection{Compared methods}

\begin{table*}[t]
\begin{center}
\caption{Quantitative results for ablation studies on model components, data modalities, QA pairs, beam width and cluster numbers. \emph{G} denotes the Generative dialog setting, while \emph{D} denotes the Discriminative one. For each group, we use the \textcolor{newgreen}{green}-\textcolor{newblue}{blue}-black color gradient to mark the scores in descending order for better visualizations.
}
\scalebox{0.96}{
\label{tab:ablation}
\begin{tabular}{|c|c|c|c|c|c|c|c|c|c|c|}
\hline
Group & Setting &  Ablation  & BLEU1  & BLEU2    & BLEU3   & BLEU4        & METEOR  & SPICE       & ROUGE\_L      & CIDEr    \\ \hline
\multirow{5}{*}{\begin{tabular}[c]{@{}c@{}}Model \\ Components \end{tabular}} & \multirow{2}{*}{\begin{tabular}[c]{@{}c@{}} G\end{tabular}} & w/o Att.  &  31.5   &   16.3  &  8.8   &  4.9  & 12.3 & 11.2 &  26.8  & 20.4    \\ \cline{3-11} 
 &  &   w/o AV-LSTM   &  32.1  &  16.2    & 8.8      &  5.1   & 12.1 & 11.3  &  27.1   &    20.3\\ \cline{2-11}
 & \multirow{3}{*}{\begin{tabular}[c]{@{}c@{}} D\end{tabular}} &  w/o Att.     &   \textcolor{newblue}{33.8}  &  \textcolor{newblue}{17.1}    &     \textcolor{newblue}{ 9.0} &  \textcolor{newblue}{5.8}   & \textcolor{newblue}{12.7}  & \textcolor{newblue}{12.3}   &    \textcolor{newblue}{ 27.6} &  \textcolor{newblue}{25.3}  \\ \cline{3-11} 
 & &  w/o AV-LSTM &  \textcolor{newblue}{32.6}    &  \textcolor{newblue}{16.8}  &8.8     & \textcolor{newblue}{5.3}      & \textcolor{newblue}{12.5} & \textcolor{newblue}{12.0}  & \textcolor{newblue}{27.5}  & \textcolor{newblue}{23.7}   \\ \cline{3-11} 
  & & w/o Reasoning& \textcolor{newgreen}{34.3} & \textcolor{newgreen}{18.3} & \textcolor{newgreen}{10.0}   & \textcolor{newgreen}{6.3}    & \textcolor{newgreen}{12.8}  & \textcolor{newgreen}{13.1} & \textcolor{newgreen}{27.9}        &     \textcolor{newgreen}{    26.3} \\ \hline \hline
  %\hline
 
\multirow{10}{*}{\begin{tabular}[c]{@{}c@{}} Data Modalities\end{tabular}}                     
& \multirow{5}{*}{\begin{tabular}[c]{@{}c@{}} G\end{tabular}} 
 & \re{w/o visual frames}        &   \textcolor{newblue}{33.1}   &  16.1   & 7.3    &    4.4   & 11.6 & 10.8  & 26.1   & 20.0    \\ \cline{3-11}
& & \re{full segmented frames}        &  \textcolor{newblue}{34.0}    &  \textcolor{newblue}{17.6}   &    \textcolor{newblue}{9.8} &  \textcolor{newblue}{6.1}     & 12.6 &  11.9 &  \textcolor{newblue}{27.7}  & 23.0     \\ \cline{3-11}
& & w/o Audio        &  32.2    &  16.2   &    \textcolor{newblue}{9.4}  &  5.4     & \textcolor{newblue}{12.8} & 11.2  &27.2    & 22.3    \\ \cline{3-11} 
& & w/o Input description    &  31.5    &  15.3   & 7.9     &  4.6     & 12.7 & 11.1  & 26.3   &    20.1 \\ \cline{3-11} 
&  & w/o HIS for \emph{A-BOT}      &  32.5    & 16.3    &    \textcolor{newblue}{9.3}  &      5.4 & 12.1 & 11.2  &27.1    & 23.0    \\ \cline{2-11} 
& \multirow{5}{*}{\begin{tabular}[c]{@{}c@{}} D\end{tabular}}
& \re{w/o segmented frames}        &    \textcolor{newblue}{33.0}  &     16.8&  9.1   &  5.9     & 12.7 & \textcolor{newblue}{12.6}  &    \textcolor{newblue}{27.8} &  \textcolor{newblue}{25.3}   \\ \cline{3-11}
& & \re{full segmented frames}        &   \textcolor{newgreen}{34.6}   &  \textcolor{newgreen}{18.8}   &    \textcolor{newgreen}{10.6} &  \textcolor{newgreen}{6.5}     &  \textcolor{newgreen}{14.0} &  \textcolor{newgreen}{13.9} &   \textcolor{newgreen}{ 28.7} &  \textcolor{newgreen}{27.7}   \\ \cline{3-11}
& &w/o Audio  &  \textcolor{newgreen}{34.6}  &    \textcolor{newblue}{18.2}  &     \textcolor{newblue}{10.2} & \textcolor{newblue}{6.2}    & \textcolor{newblue}{13.3} & \textcolor{newblue}{13.1}  &  \textcolor{newblue}{28.6}    &  \textcolor{newblue}{26.6}    \\ \cline{3-11}
 & & w/o Input description     &  32.8  &    \textcolor{newblue}{17.3}  &     8.3 &  \textcolor{newblue}{6.1}   & \textcolor{newblue}{13.1} & \textcolor{newblue}{12.8}  &  \textcolor{newblue}{27.7}    &    \textcolor{newblue}{25.8}  \\ \cline{3-11}
& & w/o HIS for \emph{A-BOT}  & \textcolor{newblue}{34.0}    &     \textcolor{newblue}{18.1} &  \textcolor{newblue}{10.5}    &  \textcolor{newblue}{6.3}   & \textcolor{newblue}{12.9} & \textcolor{newblue}{13.0}  &     \textcolor{newblue}{28.3} &     \textcolor{newblue}{26.3} \\ 
 \hline \hline

\multirow{8}{*}{\begin{tabular}[c]{@{}c@{}}QA Pairs \end{tabular}}              
&\multirow{4}{*}{\begin{tabular}[c]{@{}c@{}} G\end{tabular}}& Shuffled order  &  31.4 & 15.5     & 8.4       & 4.9     & 11.7 & 11.1  & 26.3    &  20.0    \\ \cline{3-11} 
& & Round\#2    &  27.9 &    13.3  &     7.0   &  4.0    & 10.8  & 9.7  & 24.9    &   16.7   \\ \cline{3-11}
&  & Round\#5    & 32.7  &  16.7    &   9.4     &  5.6    & 12.2 & 12.1  &  27.6   &    22.9  \\ \cline{3-11}
 &  & Round\#8   &  \textcolor{newblue}{34.1} &  \textcolor{newblue}{17.6}    &   \textcolor{newblue}{9.8}     &  \textcolor{newblue}{5.9}    & \textcolor{newblue}{12.9} & \textcolor{newblue}{12.6}  &  \textcolor{newblue}{28.5}   &   \textcolor{newblue}{ 25.5}  \\ \cline{2-11}
 &\multirow{4}{*}{\begin{tabular}[c]{@{}c@{}} D\end{tabular}}     &   Shuffled order     &   \textcolor{newblue}{34.0}        & \textcolor{newblue}{18.0}          &     \textcolor{newblue}{10.2}      &   \textcolor{newblue}{  6.1}      &  \textcolor{newblue}{12.3}  &  \textcolor{newblue}{ 13.0}   &      \textcolor{newblue}{28.1}    &  \textcolor{newblue}{ 25.4}    \\ \cline{3-11}
 & & Round\#2   & 32.2  &    17.4  &     9.0   & 5.2     & \textcolor{newblue}{12.3} & 11.9  &    27.6 &  23.0    \\ \cline{3-11}
&    & Round\#5   &  \textcolor{newblue}{34.3} &  \textcolor{newblue}{18.1}    &  \textcolor{newblue}{9.6}      & \textcolor{newblue}{5.8}     & \textcolor{newblue}{13.1} & \textcolor{newblue}{13.0}  &  \textcolor{newblue}{28.1}   &    \textcolor{newblue}{26.4}  \\ \cline{3-11}
&    & Round\#8    & \textcolor{newgreen}{35.1}  &  \textcolor{newgreen}{18.5}    &  \textcolor{newgreen}{10.4}      &  \textcolor{newgreen}{6.5}    & \textcolor{newgreen}{13.4} & \textcolor{newgreen}{13.2}  & \textcolor{newgreen}{28.8}    &  \textcolor{newgreen}{28.4}    \\ 
  \hline \hline
  
\multirow{6}{*}{\begin{tabular}[c]{@{}c@{}}Beam Width\end{tabular}}  
& \multirow{3}{*}{\begin{tabular}[c]{@{}c@{}} G\end{tabular}} &   Beam width=1     & 28.5 & 14.8 & 7.9 & 4.2 & 10.9 & 10.4 & 26.0 & 17.4 \\  \cline{3-11}
& &  Beam width=3      &  32.4 & 16.3 & 9.7 & 5.4 & \textcolor{newblue}{12.9} & 11.3 & 27.7 & 22.9 \\ \cline{3-11}
& &  Beam width=5       & 33.4 & 16.3 & \textcolor{newblue}{9.8} & 5.5 & \textcolor{newblue}{12.9} & 11.0 & 27.5 & 22.8 \\  \cline{2-11}
& \multirow{3}{*}{\begin{tabular}[c]{@{}c@{}} D\end{tabular}}&    Beam width=1    & \textcolor{newblue}{33.5} & \textcolor{newblue}{17.5} & \textcolor{newblue}{9.8} & \textcolor{newblue}{5.7} & 12.8 & \textcolor{newblue}{13.1} & \textcolor{newgreen}{28.8} & \textcolor{newblue}{24.7} \\  \cline{3-11}
&  &     Beam width=3   & \textcolor{newblue}{34.7} & \textcolor{newblue}{18.0} & \textcolor{newblue}{10.2} & \textcolor{newblue}{6.1} & \textcolor{newblue}{13.2} &\textcolor{newblue}{13.6}  & \textcolor{newblue}{28.6} & \textcolor{newgreen}{27.1} \\  \cline{3-11}
 &   &  Beam width=5     & \textcolor{newgreen}{34.9} & \textcolor{newgreen}{18.4} & \textcolor{newgreen}{10.4} & \textcolor{newgreen}{6.3} & \textcolor{newgreen}{13.3} & \textcolor{newgreen}{13.7} & \textcolor{newgreen}{28.8} & \textcolor{newgreen}{27.1} \\
\hline \hline

\multirow{3}{*}{\begin{tabular}[c]{@{}c@{}}\re{Clusters}\end{tabular}}  
& \multirow{3}{*}{\begin{tabular}[c]{@{}c@{}} D\end{tabular}} &  \re{k = 5}  & \textcolor{newblue}{34.7} & 17.7 & 9.5 & 5.8 & \textcolor{newblue}{13.1} & 13.4 & 28.3 & 26.8 \\  \cline{3-11}
& & \re{k = 10}      &  \textcolor{newblue}{34.7} & \textcolor{newblue}{18.0} & \textcolor{newgreen}{10.2} & \textcolor{newblue}{6.1} & \textcolor{newgreen}{13.2} & \textcolor{newblue}{13.6} & \textcolor{newblue}{28.6} & \textcolor{newgreen}{27.1} \\ \cline{3-11}
& &  \re{k = 15}   & \textcolor{newgreen}{34.8} & \textcolor{newgreen}{18.2} & \textcolor{newgreen}{10.2} & \textcolor{newgreen}{6.2} & \textcolor{newblue}{13.1} & \textcolor{newgreen}{13.7} & \textcolor{newgreen}{28.8} & \textcolor{newblue}{27.0} \\  \cline{2-11}
\hline
\end{tabular}}
\end{center}
\end{table*}

We categorize the experiments into multiple groups to provide a more comprehensive and objective analysis for the unseen video description task and the proposed methods.

\noindent \textbf{A-BOT for Description.} 
To better understand and illustrate the knowledge gap as well as the transfer process between two dialog agents, we include the experiments for \emph{A-BOT} to accomplish the same video description task. In this case, our video description task can be also considered as a classic video captioning task. \emph{A-BOT} has access to the video data (\textit{i.e.}, the visual frames and audio signals), and is asked to describe the video. The dialog history and the original input descriptions are removed from the input for \emph{A-BOT} to reduce bias for this group of experiments. 

\noindent \textbf{Basic Baselines.}
The basic baselines are obtained without the cooperative internal dialog process, which means that the \emph{Q-BOT} is asked to directly describe the video based on the existing dialog history input without additional chances to ask questions. In this case, the number of testing cases remains to be 7330.

\noindent \textbf{Strong Baselines.}
The strong baselines are established by providing the full 10 rounds \textit{ground truth dialog} to \emph{Q-BOT}. \emph{Q-BOT} can thus directly generate the final descriptions without asking questions. It can be regarded as an "upper bound case" for the generative dialog setting to some extent, since the internal dialog is trained to imitate the ground truth QA interactions. 

\noindent \textbf{Other Baselines.}
We also investigate the performance using the previously proposed methods~\cite{hori2019end,schwartz2019simple}. However, the previous methods are initially designed for question answering tasks, therefore, we modify the generators to generate video descriptions after 10 rounds of QA interactions and fine-tune the models for our task. All the data input for these baselines remains the same as for the proposed QA-Cooperative networks.

\noindent \re{\textbf{Simulated Human Evaluation.}
Considering the intended practical scenario for our proposed task involves the interactions between AI systems (\textit{i.e.}, the role of \emph{Q-BOT} in this work) and the real human users (\textit{i.e.}, the role of \emph{A-BOT} in this work), we also perform a simulated human evaluation test for the discriminative dialog setting. During the simulated test, the answers given by \emph{A-BOT} are replaced by the ground truth answers correspond to the picked question by \emph{Q-BOT} in inference. The training process remains the same as previously described in Section~\ref{subsec:learning}.
}

\subsection{Video Description Results}
The quantitative experimental results for the final description are shown in Table~\ref{tab:results}. We observe that \emph{A-BOT} performs better than \emph{Q-BOT} as expected. 
% It is reasonable due to the fact that \emph{A-BOT} has direct access to the original video data, and the performance difference between two agents also experimentally demonstrates the actual knowledge gap due to the lack of direct information sources. 
However, with the proposed QA-Cooperative networks and cooperative learning methods, our \emph{Q-BOT} can achieve very promising performance under both generative and discriminative internal dialog settings, especially with the discriminative setting. 
\re{In the meanwhile, the above observations also show that although the dialog has been proven to be an effective information source to supplement the incomplete visual data compared to the basic baseline setting without dialog, it is rather difficult to completely alternate the missing visual information. The results are also consistent with the knowledge gap observed from two types of descriptions from the dataset as described in Section~\ref{sec:introduction} due to the lack of visual data}.

 We also present the experimental results obtained under the initial task setup from~\cite{yzhu2020describing}, where the visual input for \emph{Q-BOT} is two original static RGB frames from the video without semantic segmentation. The performance shows no evident gap between the two task setups, demonstrating that our models extract the useful information from the dialog history, instead of benefiting from the possible bias introduced in the visual input. It is worth noting that this observation is not contradictory to the previous statement about the significance of the visual data. The previous one emphasizes its importance from the existence and temporal aspect, while the raw/segmented visual information addresses more on the details within the same frame.

The improvement for the final description performance compared to basic baselines shows the effectiveness of the knowledge transfer process between two agents with unbalanced input data. For the generative internal dialog setting, our \emph{Q-BOT} with the QA-Cooperative network can achieve comparable performance close to the strong baselines where the full ground truth dialog is provided. In contrast, for the discriminative setting, our \emph{Q-BOT} is able to outperform the strong baselines for most of the evaluation metrics, the primary metric CIDEr score achieves 27.1. 
The simulated human evaluation yields better performance compared to the case with both dialog agents.
In addition, we also notice that the MM attention mechanism helps with performance improvement compared to the IM attention mechanism.

\begin{figure*}[t]
    \centering
    \includegraphics[width=0.98\textwidth]{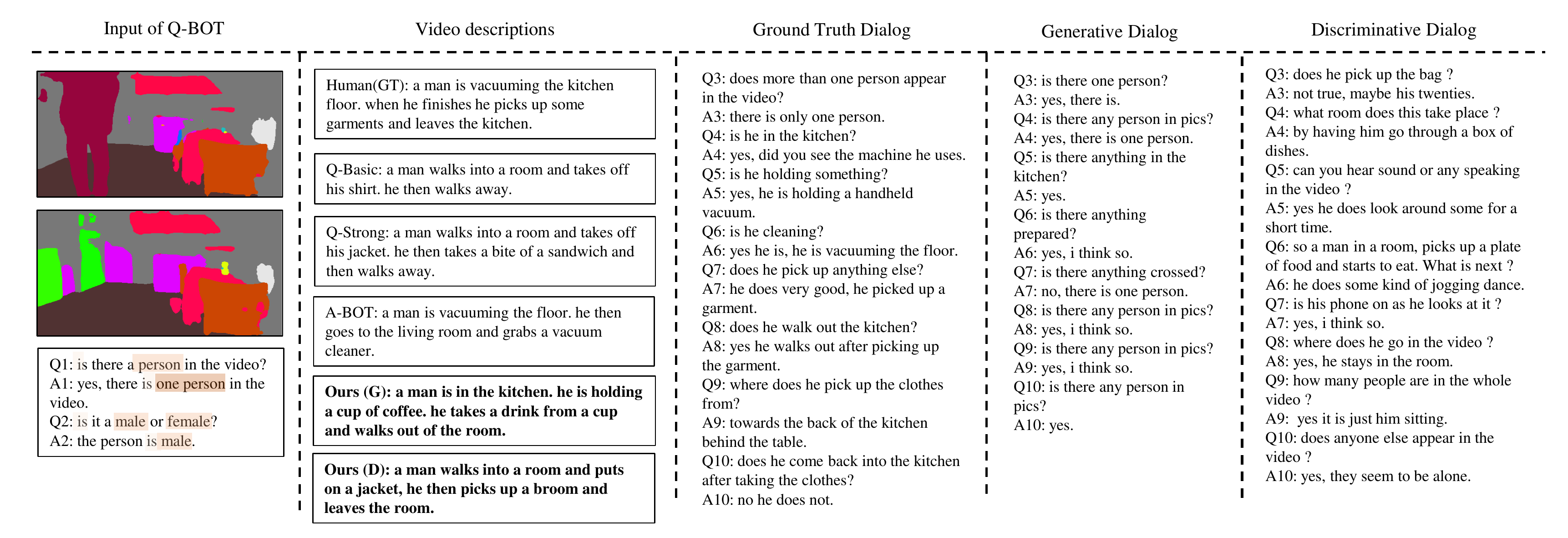}
    \caption{Example of qualitative results. We present the input of Q-BOT, different video descriptions, and the internal dialog.
    The descriptions given by our \emph{Q-BOT} include more details compared to multiple baselines. The color intensities in the figure represent attention weights. More examples can be found in Appendix.}
    \label{fig4:qualitative}
\end{figure*}

Figure~\ref{fig4:qualitative} shows examples of qualitative results.
Due to the limited space, more qualitative examples can be found in Appendix.
The qualitative examples reveal the consistent results with our quantitative evaluations, the video descriptions generated by \emph{Q-BOT} with our proposed QA-Cooperative networks contain more detailed information compared to the basic baselines and are more close to the strong baseline cases where the full ground truth dialog is provided as input. 
The examples in Figure~\ref{fig4:qualitative} is challenging test cases due to the fact that only a few rounds of QA pairs are included in the input, however, the final descriptions given by our \emph{Q-BOT} contains the concrete information such as the room types (\textit{e.g.}, the kitchen) that are not included in the input. It demonstrates that our \emph{Q-BOT} does benefit from the effective knowledge transfer process via the natural language dialog to describe the unseen videos.
We also notice from the qualitative results that the internal dialog obtained under the generative internal dialog setting tends to contain repetitive information, which is also observed from previous work on the dialog agents~\cite{das2017learning}. As for comparisons, the questions and answers selected under the discriminative internal dialog setting are more diverse and informative, which explains the reason for the better final descriptions. 
% Due to the limited space, more qualitative examples can be found in Appendix.

\subsection{Ablation Studies}

We continue to conduct extensive ablation studies on model components, data modalities, QA pairs, and beam width in this section to better analyze the proposed methods. Note that all the experiments in this section adopt the MM attention module as the attention mechanism since the MM attention module is proved to be more effective than IM attention module in previous experiments in Table~\ref{tab:results}.

\subsubsection{Model Components}

\noindent \textbf{Attention Modules.} 
We propose two attention mechanisms in our QA-Cooperative network architectures, \textit{i.e.}, the MM (Multi-Modal) attention module and the IM (Intra-Module) attention module. 
Interestingly, we observe that both attention mechanisms help to improve the final performance of the video description, which is different from the results in~\cite{schwartz2019simple}.  In~\cite{schwartz2019simple}, the authors find that the attention on the dialog history does not yield performance improvements for the AVSD task. One possible reason for this difference could probably be explained from the perspective of causal inference as in~\cite{qi2019two}, where the dialog history is found to be a spurious and biased factor and should be removed for classic question answering tasks. This again emphasizes the difference of our task from the classic VQA and visual dialog tasks from a novel angle of causality. The dialog history for \emph{Q-BOT} in our unseen video description task is not a spurious factor but a significant information source. The attention modules help to raise the CIDEr scores from 20.4 to 22.9 for the generative setting, and from 25.3 to 27.1 for the discriminative setting.

\noindent \textbf{Audio-Visual LSTM.}
The Audio-Visual LSTM component is designed for \emph{A-BOT} to process the audio and visual input data in addition to the MM module. The output of this model component is used as the initial state input for \emph{A-BOT}. We test the ablation experiments by removing this component. Experiments show that the audio-visual LSTM contributes to the final description performance. This module accounts for a raise of 2.6 and 3.4 on CIDEr scores for the generative and discriminative settings, respectively.

\noindent \textbf{Internal Selection.}
Under the discriminative setting, we address the internal selective abilities of two agents by adopting the sparse annotations as explained in Section~\ref{subsec:reasoning}. This ablation experiments prove the contributions of this module, leading to an increase of 0.8 for the CIDEr score.

% In addition to the final description performance in \emph{w/o Reasoning} setting of Table~\ref{tab:ablation}, we also calculate the ground truth question and answer selection ratio as qualitative evaluation as shown in Figure~\ref{fig3:reasoning}. 
% The ground truth selection ratios increase after deploying the internal reasoning mechanism. Additionally, we also observe that the selection ratio for questions is generally higher than the ratio for answers.
% The selection ratios also increase as more ground truth QA pairs are provided as input (\textit{i.e.}, with larger starting round number), as in Figure~\ref{fig3:reasoning}, the ground truth selection ratios with the starting round number 8 are generally higher than the starting round number 2.

% \begin{figure}[t]
%     \centering
%     \includegraphics[width=0.48\textwidth]{fig/ratio.png}
%     \caption{Ground truth question and answer selection ratio during training. We plot the selection ratios with the starting round number 2 and 8 as examples. The solid lines represent the ground truth selection ratios for questions, the dotted lines are the ratios for answers.}
%     \label{fig3:reasoning}
% \end{figure}

\subsubsection{Data Modalities}

% In this work, we propose to demonstrate the feasibility and effectiveness to supplement the incomplete visual data with the natural language dialog.
Our unseen video description task incorporates multiple modalities of data, we therefore perform the ablation studies to analyze the impact of different data modalities. 

\noindent \re{\textbf{Visual Data.}}
\re{In our experimental settings, \emph{Q-BOT} takes two segmented visual frames as the implicit visual input. We exploit the impact of different types of visual input on the final video description performance. Specifically, we conduct ablation studies with the full segmented frames and without any visual frames under both dialog settings. The experimental results demonstrate the significance of the visual data for our proposed task. It is worth noting that even the visual input are processed with segmentation operation, we obverse performance improve given more video frames. In contrast, the complete removal of visual data from input causes the relatively poor performance for video descriptions. 
}

\noindent \textbf{Audio Data.}
Audio data forms part of the input for \emph{A-BOT}, since the audio perception is another important information source in addition to the vision for humans.
We remove the audio data from the input for \emph{A-BOT} to investigate its influence on the final description performance. We observe from Table~\ref{tab:ablation} that audio data contributes to the better final description performance under both internal dialog settings.

\noindent \textbf{Input Description Data.}
The input descriptions obtained from the human annotators are also provided to \emph{A-BOT} under our task setup.
We study the performance of final descriptions after removing the input description data from the input for \emph{A-BOT}.
Table~\ref{tab:ablation} shows that the input descriptions also contribute to the final description performance.

\noindent \textbf{Dialog History.}
The dialog history is a common input for both \emph{Q-BOT} and \emph{A-BOT}. It is the major information source for \emph{Q-BOT} to describe the unseen video, which is already demonstrated in the first basic baseline situation in Table~\ref{tab:results}.
Therefore, for the ablation studies, we conduct experiments under the situation where the dialog history is invisible to \emph{A-BOT}. Interestingly, the performance is not much impaired in this case compared to other data modalities. Intuitively, it is also reasonable due to the fact that \emph{A-BOT} does not rely on the dialog history to provide answers to the questions raised by \emph{Q-BOT}, since \emph{A-BOT} has already watched the entire video. This finding is also consistent with the previous findings for classic question answering tasks in~\cite{schwartz2019simple} and~\cite{qi2019two}. In~\cite{schwartz2019simple}, the authors find that the attention on the dialog history does not yield performance improvement for answering questions. In the work of Qi \textit{et al.}~\cite{qi2019two}, the dialog history is proven to be a spurious factor that ultimately impairs the performance for question answering tasks.

\subsubsection{QA Pairs}
\label{subsec:qapairs}

\noindent \textbf{Order of QA Pairs.} We test the impact of the order of the QA pairs by randomly shuffle the orders in the dialog history. Similar to the observations from~\cite{alamri2019audio,yzhu2020describing}, the QA in the dialog history order is an important factor that influences the performance of the final descriptions. With the shuffled dialog history, the performance is impacted under both generative and discriminative dialog settings. We observe that the primary CIDEr scores drop 2.9 and 1.7 for the generative and discriminative settings, respectively.

\noindent \textbf{Number of Input QA Pairs.} 
We also take a closer look at the experimental results with different numbers of QA pairs included in the input dialog history . In other words, we modify the number of starting round for the testing cases. 
Unsurprisingly, the more ground-truth QA pairs in the dialog history usually lead to the better final performance for describing the unseen videos.

\subsubsection{Other Hyper-parameters}

\noindent \textbf{Beam Search.}
We use the beam search when generating the final descriptions. We experiment with different numbers of beam width. The experimental results from Table~\ref{tab:ablation} shows that with wider beam width, the final performance for the unseen video description tasks improves. However, a beam width of 3 is generally adequate for achieving good results. For the main experimental results reported in Table~\ref{tab:results}, we adopt the beam width of 3.

\noindent \textbf{Number of Clusters.} In the discriminative internal dialog setting, \emph{Q-BOT} selects the questions following a two-phase selection mechanism.
The question candidates are firstly processed using unsupervised k-means clusters algorithm. We test different numbers of clusters to study its impact. The experimental results show that a larger number of clusters leads to slightly better performance. In our main experiments, we use 10 clusters in the first selection phase.

\section{Conclusion and Discussion}

In this work, we propose a novel multi-modal task that aims to describe an unseen video based on the incomplete visual input and the natural language dialog. 
There are two primary motivations behind this work: to introduce a more reliable task setup by providing AI with implicit visual input, and to demonstrate the effectiveness of using the natural language dialog as the additional source to supplement the missing visual information.
We propose two different experimental settings with their corresponding cooperative network models that effectively help with the knowledge transfer process between two agents. Extensive experiments demonstrate the promising and competitive performance of the proposed methods over multiple baselines. 

There are research directions that worth further exploiting in the future: (a) One possible direction could be encouraging more efficient dialog interactions between two agents. Specifically, we observe from the experiments that the dialog agent \emph{Q-BOT} does not always need 10 question chances to achieve good performance for the video descriptions. It may already have enough information to summarize the video at the end of eight or nine rounds of QA interactions. It would be therefore interesting to further encourage more efficient information exchange, and to exploit the possible early stop mechanism for the dialog interactions. 
\re{(b) The simulated human evaluation results indicate that there is still room to enhance \emph{A-BOT}'s ability to achieve better performance for the proposed video description task.}
(c) It would also be interesting to further refine the task formulation and to design more specific ultimate objectives other than general video descriptions, \textit{e.g.}, we could ask \emph{Q-BOT} to generate a scene graph mainly based on the natural language dialog. (d) A more sophisticated mechanism that enables \emph{Q-BOT} to ask guided and structured questions could also be useful when applied in real-life scenarios.

\section{Acknowledgements}
This research was partially supported by NSF NeTS-2109982 and the gift donation from Cisco. This article solely reflects the opinions and conclusions of its authors and not the funding agents.

\ifCLASSOPTIONcompsoc
  %\section*{Acknowledgments}
\else
  %\section*{Acknowledgment}
\fi

%This work is supported by

\ifCLASSOPTIONcaptionsoff
  \newpage
\fi



%\bibliographystyle{IEEEtran}

{
\bibliographystyle{ref.bst}
\bibliography{egbib}

\begin{thebibliography}{10}\itemsep=-1pt

\bibitem{agarwal2020history}
Shubham Agarwal, Trung Bui, Joon-Young Lee, Ioannis Konstas, and Verena Rieser.
\newblock History for visual dialog: Do we really need it?
\newblock In {\em ACL}, 2020.

\bibitem{Agarwal_2020_CVPR}
Vedika Agarwal, Rakshith Shetty, and Mario Fritz.
\newblock Towards causal vqa: Revealing and reducing spurious correlations by
  invariant and covariant semantic editing.
\newblock In {\em CVPR}, 2020.

\bibitem{alamri2019audio}
Huda Alamri, Vincent Cartillier, Abhishek Das, Jue Wang, Anoop Cherian, Irfan
  Essa, Dhruv Batra, Tim~K Marks, Chiori Hori, Peter Anderson, et~al.
\newblock Audio visual scene-aware dialog.
\newblock In {\em CVPR}, 2019.

\bibitem{anderson2016spice}
Peter Anderson, Basura Fernando, Mark Johnson, and Stephen Gould.
\newblock Spice: Semantic propositional image caption evaluation.
\newblock In {\em ECCV}, 2016.

\bibitem{anderson2018bottom}
Peter Anderson, Xiaodong He, Chris Buehler, Damien Teney, Mark Johnson, Stephen
  Gould, and Lei Zhang.
\newblock Bottom-up and top-down attention for image captioning and visual
  question answering.
\newblock In {\em CVPR}, 2018.

\bibitem{antol2015vqa}
Stanislaw Antol, Aishwarya Agrawal, Jiasen Lu, Margaret Mitchell, Dhruv Batra,
  C Lawrence~Zitnick, and Devi Parikh.
\newblock Vqa: Visual question answering.
\newblock In {\em ICCV}, 2015.

\bibitem{arandjelovic2018objects}
Relja Arandjelovic and Andrew Zisserman.
\newblock Objects that sound.
\newblock In {\em ECCV}, 2018.

\bibitem{banerjee2005meteor}
Satanjeev Banerjee and Alon Lavie.
\newblock Meteor: An automatic metric for mt evaluation with improved
  correlation with human judgments.
\newblock In {\em Proceedings of the acl workshop on intrinsic and extrinsic
  evaluation measures for machine translation and/or summarization}, pages
  65--72, 2005.

\bibitem{chen2019variational}
Fuhai Chen, Rongrong Ji, Jiayi Ji, Xiaoshuai Sun, Baochang Zhang, Xuri Ge,
  Yongjian Wu, Feiyue Huang, and Yan Wang.
\newblock Variational structured semantic inference for diverse image
  captioning.
\newblock In {\em NeurIPS}, 2019.

\bibitem{chen2020counterfactual}
Long Chen, Xin Yan, Jun Xiao, Hanwang Zhang, Shiliang Pu, and Yueting Zhuang.
\newblock Counterfactual samples synthesizing for robust visual question
  answering.
\newblock In {\em CVPR}, 2020.

\bibitem{chen2017sca}
Long Chen, Hanwang Zhang, Jun Xiao, Liqiang Nie, Jian Shao, Wei Liu, and
  Tat-Seng Chua.
\newblock Sca-cnn: Spatial and channel-wise attention in convolutional networks
  for image captioning.
\newblock In {\em CVPR}, 2017.

\bibitem{Cornia_2020_CVPR}
Marcella Cornia, Matteo Stefanini, Lorenzo Baraldi, and Rita Cucchiara.
\newblock Meshed-memory transformer for image captioning.
\newblock In {\em CVPR}, 2020.

\bibitem{das2017human}
Abhishek Das, Harsh Agrawal, Larry Zitnick, Devi Parikh, and Dhruv Batra.
\newblock Human attention in visual question answering: Do humans and deep
  networks look at the same regions?
\newblock {\em Computer Vision and Image Understanding}, 163:90--100, 2017.

\bibitem{das2017visual}
Abhishek Das, Satwik Kottur, Khushi Gupta, Avi Singh, Deshraj Yadav,
  Jos{\'e}~MF Moura, Devi Parikh, and Dhruv Batra.
\newblock Visual dialog.
\newblock In {\em CVPR}, 2017.

\bibitem{das2017learning}
Abhishek Das, Satwik Kottur, Jos{\'e}~MF Moura, Stefan Lee, and Dhruv Batra.
\newblock Learning cooperative visual dialog agents with deep reinforcement
  learning.
\newblock In {\em ICCV}, 2017.

\bibitem{de2017guesswhat}
Harm De~Vries, Florian Strub, Sarath Chandar, Olivier Pietquin, Hugo
  Larochelle, and Aaron Courville.
\newblock Guesswhat?! visual object discovery through multi-modal dialogue.
\newblock In {\em CVPR}, 2017.

\bibitem{devlin2018bert}
Jacob Devlin, Ming-Wei Chang, Kenton Lee, and Kristina Toutanova.
\newblock Bert: Pre-training of deep bidirectional transformers for language
  understanding.
\newblock {\em arXiv preprint arXiv:1810.04805}, 2018.

\bibitem{duan2020audio}
Bin Duan, Hao Tang, Wei Wang, Ziliang Zong, Guowei Yang, and Yan Yan.
\newblock Audio-visual event localization via recursive fusion by joint
  co-attention.
\newblock In {\em WACV}, 2021.

\bibitem{gan2019multi}
Zhe Gan, Yu Cheng, Ahmed Kholy, Linjie Li, Jingjing Liu, and Jianfeng Gao.
\newblock Multi-step reasoning via recurrent dual attention for visual dialog.
\newblock In {\em ACL}, 2019.

\bibitem{gao2018learning}
Ruohan Gao, Rogerio Feris, and Kristen Grauman.
\newblock Learning to separate object sounds by watching unlabeled video.
\newblock In {\em ECCV}, 2018.

\bibitem{Gao_2020_CVPR}
Ruohan Gao, Tae-Hyun Oh, Kristen Grauman, and Lorenzo Torresani.
\newblock Listen to look: Action recognition by previewing audio.
\newblock In {\em CVPR}, 2020.

\bibitem{Guo_2020_CVPR}
Dan Guo, Hui Wang, Hanwang Zhang, Zheng-Jun Zha, and Meng Wang.
\newblock Iterative context-aware graph inference for visual dialog.
\newblock In {\em Proceedings of the IEEE/CVF Conference on Computer Vision and
  Pattern Recognition (CVPR)}, June 2020.

\bibitem{Guo_normalized}
Longteng Guo, Jing Liu, Xinxin Zhu, Peng Yao, Shichen Lu, and Hanqing Lu.
\newblock Normalized and geometry-aware self-attention network for image
  captioning.
\newblock In {\em CVPR}, 2020.

\bibitem{guo2018dialog}
Xiaoxiao Guo, Hui Wu, Yu Cheng, Steven Rennie, Gerald Tesauro, and
  Rogerio~Schmidt Feris.
\newblock Dialog-based interactive image retrieval.
\newblock In {\em NeurIPS}, 2018.

\bibitem{hershey2017cnn}
Shawn Hershey, Sourish Chaudhuri, Daniel~PW Ellis, Jort~F Gemmeke, Aren Jansen,
  R~Channing Moore, Manoj Plakal, Devin Platt, Rif~A Saurous, Bryan Seybold,
  et~al.
\newblock Cnn architectures for large-scale audio classification.
\newblock In {\em ICASSP}. IEEE, 2017.

\bibitem{hochreiter1997long}
Sepp Hochreiter and J{\"u}rgen Schmidhuber.
\newblock Long short-term memory.
\newblock {\em Neural computation}, 1997.

\bibitem{hori2019end}
Chiori Hori, Huda Alamri, Jue Wang, Gordon Wichern, Takaaki Hori, Anoop
  Cherian, Tim~K Marks, Vincent Cartillier, Raphael~Gontijo Lopes, Abhishek
  Das, et~al.
\newblock End-to-end audio visual scene-aware dialog using multimodal
  attention-based video features.
\newblock In {\em ICASSP}. IEEE, 2019.

\bibitem{hu2019deep}
Di Hu, Feiping Nie, and Xuelong Li.
\newblock Deep multimodal clustering for unsupervised audiovisual learning.
\newblock In {\em CVPR}, 2019.

\bibitem{jain2018two}
Unnat Jain, Svetlana Lazebnik, and Alexander~G Schwing.
\newblock Two can play this game: visual dialog with discriminative question
  generation and answering.
\newblock In {\em CVPR}, 2018.

\bibitem{kastner1999increased}
Sabine Kastner, Mark~A Pinsk, Peter De~Weerd, Robert Desimone, and Leslie~G
  Ungerleider.
\newblock Increased activity in human visual cortex during directed attention
  in the absence of visual stimulation.
\newblock {\em Neuron}, 1999.

\bibitem{lee2018answerer}
Sang-Woo Lee, Yu-Jung Heo, and Byoung-Tak Zhang.
\newblock Answerer in questioner's mind: information theoretic approach to
  goal-oriented visual dialog.
\newblock In {\em NeurIPS}, 2018.

\bibitem{lin2004rouge}
Chin-Yew Lin.
\newblock Rouge: A package for automatic evaluation of summaries.
\newblock In {\em Text summarization branches out}, pages 74--81, 2004.

\bibitem{lu2017knowing}
Jiasen Lu, Caiming Xiong, Devi Parikh, and Richard Socher.
\newblock Knowing when to look: Adaptive attention via a visual sentinel for
  image captioning.
\newblock In {\em CVPR}, 2017.

\bibitem{lu2016hierarchical}
Jiasen Lu, Jianwei Yang, Dhruv Batra, and Devi Parikh.
\newblock Hierarchical question-image co-attention for visual question
  answering.
\newblock In {\em NeurIPS}, 2016.

\bibitem{massiceti2018flipdial}
Daniela Massiceti, N Siddharth, Puneet~K Dokania, and Philip~HS Torr.
\newblock Flipdial: A generative model for two-way visual dialogue.
\newblock In {\em CVPR}, 2018.

\bibitem{murahari2020large}
Vishvak Murahari, Dhruv Batra, Devi Parikh, and Abhishek Das.
\newblock Large-scale pretraining for visual dialog: A simple state-of-the-art
  baseline.
\newblock In {\em ECCV}. Springer, 2020.

\bibitem{owens2018audio}
Andrew Owens and Alexei~A Efros.
\newblock Audio-visual scene analysis with self-supervised multisensory
  features.
\newblock In {\em ECCV}, 2018.

\bibitem{owens2018learning}
Andrew Owens, Jiajun Wu, Josh~H McDermott, William~T Freeman, and Antonio
  Torralba.
\newblock Learning sight from sound: Ambient sound provides supervision for
  visual learning.
\newblock {\em International Journal of Computer Vision}, 126(10):1120--1137,
  2018.

\bibitem{pan2017video}
Yingwei Pan, Ting Yao, Houqiang Li, and Tao Mei.
\newblock Video captioning with transferred semantic attributes.
\newblock In {\em CVPR}, 2017.

\bibitem{Pan_2020_CVPR}
Yingwei Pan, Ting Yao, Yehao Li, and Tao Mei.
\newblock X-linear attention networks for image captioning.
\newblock In {\em CVPR}, 2020.

\bibitem{papineni2002bleu}
Kishore Papineni, Salim Roukos, Todd Ward, and Wei-Jing Zhu.
\newblock Bleu: a method for automatic evaluation of machine translation.
\newblock In {\em ACL}, 2002.

\bibitem{pennington2014glove}
Jeffrey Pennington, Richard Socher, and Christopher~D Manning.
\newblock Glove: Global vectors for word representation.
\newblock In {\em EMNLP}, 2014.

\bibitem{qi2019two}
Jiaxin Qi, Yulei Niu, Jianqiang Huang, and Hanwang Zhang.
\newblock Two causal principles for improving visual dialog.
\newblock In {\em CVPR}, 2020.

\bibitem{rajendran2018learning}
Janarthanan Rajendran, Jatin Ganhotra, Satinder Singh, and Lazaros Polymenakos.
\newblock Learning end-to-end goal-oriented dialog with multiple answers.
\newblock In {\em EMNLP}, 2018.

\bibitem{rennie2017self}
Steven~J Rennie, Etienne Marcheret, Youssef Mroueh, Jerret Ross, and Vaibhava
  Goel.
\newblock Self-critical sequence training for image captioning.
\newblock In {\em CVPR}, 2017.

\bibitem{rouditchenko2019self}
Andrew Rouditchenko, Hang Zhao, Chuang Gan, Josh McDermott, and Antonio
  Torralba.
\newblock Self-supervised audio-visual co-segmentation.
\newblock In {\em ICASSP}. IEEE, 2019.

\bibitem{roy2000spoken}
Nicholas Roy, Joelle Pineau, and Sebastian Thrun.
\newblock Spoken dialogue management using probabilistic reasoning.
\newblock In {\em ACL}, 2000.

\bibitem{schwartz2019simple}
Idan Schwartz, Alexander~G Schwing, and Tamir Hazan.
\newblock A simple baseline for audio-visual scene-aware dialog.
\newblock In {\em CVPR}, 2019.

\bibitem{schwartz2019factor}
Idan Schwartz, Seunghak Yu, Tamir Hazan, and Alexander~G Schwing.
\newblock Factor graph attention.
\newblock In {\em CVPR}, 2019.

\bibitem{senocak2018learning}
Arda Senocak, Tae-Hyun Oh, Junsik Kim, Ming-Hsuan Yang, and In So~Kweon.
\newblock Learning to localize sound source in visual scenes.
\newblock In {\em CVPR}, 2018.

\bibitem{seo2017visual}
Paul~Hongsuck Seo, Andreas Lehrmann, Bohyung Han, and Leonid Sigal.
\newblock Visual reference resolution using attention memory for visual dialog.
\newblock In {\em NeurIPS}, 2017.

\bibitem{shih2016look}
Kevin~J Shih, Saurabh Singh, and Derek Hoiem.
\newblock Where to look: Focus regions for visual question answering.
\newblock In {\em CVPR}, 2016.

\bibitem{shukla2019should}
Pushkar Shukla, Carlos Elmadjian, Richika Sharan, Vivek Kulkarni, Matthew Turk,
  and William~Yang Wang.
\newblock What should i ask? using conversationally informative rewards for
  goal-oriented visual dialog.
\newblock In {\em ACL}, 2019.

\bibitem{singh1999reinforcement}
Satinder~P Singh, Michael~J Kearns, Diane~J Litman, and Marilyn~A Walker.
\newblock Reinforcement learning for spoken dialogue systems.
\newblock In {\em Nips}, 1999.

\bibitem{song2018explore}
Xiaomeng Song, Yucheng Shi, Xin Chen, and Yahong Han.
\newblock Explore multi-step reasoning in video question answering.
\newblock In {\em ACM Multimedia}, 2018.

\bibitem{strub2017end}
Florian Strub, Harm De~Vries, J{\'e}r{\'e}mie Mary, Bilal Piot, Aaron
  Courville, and Olivier Pietquin.
\newblock End-to-end optimization of goal-driven and visually grounded dialogue
  systems.
\newblock In {\em IJCAI}, 2017.

\bibitem{tian2018audio}
Yapeng Tian, Jing Shi, Bochen Li, Zhiyao Duan, and Chenliang Xu.
\newblock Audio-visual event localization in unconstrained videos.
\newblock In {\em ECCV}, 2018.

\bibitem{ungerleider2000mechanisms}
Sabine~Kastner Ungerleider and Leslie G.
\newblock Mechanisms of visual attention in the human cortex.
\newblock {\em Annual review of neuroscience}, 2000.

\bibitem{vedantam2015cider}
Ramakrishna Vedantam, C Lawrence~Zitnick, and Devi Parikh.
\newblock Cider: Consensus-based image description evaluation.
\newblock In {\em CVPR}, 2015.

\bibitem{wang2018reconstruction}
Bairui Wang, Lin Ma, Wei Zhang, and Wei Liu.
\newblock Reconstruction network for video captioning.
\newblock In {\em CVPR}, 2018.

\bibitem{wang2018bidirectional}
Jingwen Wang, Wenhao Jiang, Lin Ma, Wei Liu, and Yong Xu.
\newblock Bidirectional attentive fusion with context gating for dense video
  captioning.
\newblock In {\em CVPR}, 2018.

\bibitem{wang2017diverse}
Liwei Wang, Alexander Schwing, and Svetlana Lazebnik.
\newblock Diverse and accurate image description using a variational
  auto-encoder with an additive gaussian encoding space.
\newblock In {\em NeurIPS}, 2017.

\bibitem{wang2020vd}
Yue Wang, Shafiq Joty, Michael~R Lyu, Irwin King, Caiming Xiong, and Steven~CH
  Hoi.
\newblock Vd-bert: A unified vision and dialog transformer with bert.
\newblock In {\em EMNLP}, 2020.

\bibitem{wu2018you}
Qi Wu, Peng Wang, Chunhua Shen, Ian Reid, and Anton Van Den~Hengel.
\newblock Are you talking to me? reasoned visual dialog generation through
  adversarial learning.
\newblock In {\em CVPR}, 2018.

\bibitem{wu2020revisiting}
Yu Wu, Lu Jiang, and Yi Yang.
\newblock Revisiting embodiedqa: A simple baseline and beyond.
\newblock {\em IEEE Transactions on Image Processing}, 29:3984--3992, 2020.

\bibitem{wu2021explore}
Yu Wu and Yi Yang.
\newblock Exploring heterogeneous clues for weakly-supervised audio-visual
  video parsing.
\newblock In {\em CVPR}, 2021.

\bibitem{wu2018decoupled}
Yu Wu, Linchao Zhu, Lu Jiang, and Yi Yang.
\newblock Decoupled novel object captioner.
\newblock In {\em ACM Multimedia}, 2018.

\bibitem{wu2019dual}
Yu Wu, Linchao Zhu, Yan Yan, and Yi Yang.
\newblock Dual attention matching for audio-visual event localization.
\newblock In {\em ICCV}, 2019.

\bibitem{xiong2016dynamic}
Caiming Xiong, Stephen Merity, and Richard Socher.
\newblock Dynamic memory networks for visual and textual question answering.
\newblock In {\em ICML}, 2016.

\bibitem{xu2016ask}
Huijuan Xu and Kate Saenko.
\newblock Ask, attend and answer: Exploring question-guided spatial attention
  for visual question answering.
\newblock In {\em ECCV}. Springer, 2016.

\bibitem{xu2015show}
Kelvin Xu, Jimmy Ba, Ryan Kiros, Kyunghyun Cho, Aaron Courville, Ruslan
  Salakhudinov, Rich Zemel, and Yoshua Bengio.
\newblock Show, attend and tell: Neural image caption generation with visual
  attention.
\newblock In {\em ICML}, 2015.

\bibitem{yang2016stacked}
Zichao Yang, Xiaodong He, Jianfeng Gao, Li Deng, and Alex Smola.
\newblock Stacked attention networks for image question answering.
\newblock In {\em CVPR}, 2016.

\bibitem{yao2017boosting}
Ting Yao, Yingwei Pan, Yehao Li, Zhaofan Qiu, and Tao Mei.
\newblock Boosting image captioning with attributes.
\newblock In {\em ICCV}, 2017.

\bibitem{you2016image}
Quanzeng You, Hailin Jin, Zhaowen Wang, Chen Fang, and Jiebo Luo.
\newblock Image captioning with semantic attention.
\newblock In {\em CVPR}, 2016.

\bibitem{zhang2018goal}
Junjie Zhang, Qi Wu, Chunhua Shen, Jian Zhang, Jianfeng Lu, and Anton Van
  Den~Hengel.
\newblock Goal-oriented visual question generation via intermediate rewards.
\newblock In {\em ECCV}, 2018.

\bibitem{zhao2018sound}
Hang Zhao, Chuang Gan, Andrew Rouditchenko, Carl Vondrick, Josh McDermott, and
  Antonio Torralba.
\newblock The sound of pixels.
\newblock In {\em ECCV}, 2018.

\bibitem{zhou2017scene}
Bolei Zhou, Hang Zhao, Xavier Puig, Sanja Fidler, Adela Barriuso, and Antonio
  Torralba.
\newblock Scene parsing through ade20k dataset.
\newblock In {\em CVPR}, 2017.

\bibitem{zhou2018semantic}
Bolei Zhou, Hang Zhao, Xavier Puig, Tete Xiao, Sanja Fidler, Adela Barriuso,
  and Antonio Torralba.
\newblock Semantic understanding of scenes through the ade20k dataset.
\newblock {\em IJCV}, 2018.

\bibitem{zhou2020unified}
Luowei Zhou, Hamid Palangi, Lei Zhang, Houdong Hu, Jason Corso, and Jianfeng
  Gao.
\newblock Unified vision-language pre-training for image captioning and vqa.
\newblock In {\em AAAI}, 2020.

\bibitem{zhou2018end}
Luowei Zhou, Yingbo Zhou, Jason~J Corso, Richard Socher, and Caiming Xiong.
\newblock End-to-end dense video captioning with masked transformer.
\newblock In {\em CVPR}, 2018.

\bibitem{zhu2021learning}
Ye Zhu, Yu Wu, Hugo Latapie, Yi Yang, and Yan Yan.
\newblock Learning audio-visual correlations from variational cross-modal
  generation.
\newblock In {\em ICCASP}, 2021.

\bibitem{yzhu2020describing}
Ye Zhu, Yu Wu, Yi Yang, and Yan Yan.
\newblock Describing unseen videos via multi-modal cooperative dialog agents.
\newblock In {\em ECCV}, 2020.

\end{thebibliography}
}

\newpage
% \newpage
% \newpage
\appendices

\section{Knowledge Gap and Two Types of Descriptions from the AVSD Datasets}

Our proposed task involves two different types of video descriptions as mentioned in the Introduction. In this Appendix, we provide further clarifications on the differences between the input and final descriptions, which helps to better understand the concept of knowledge gap caused by the implicit visions.

Table~\ref{tab4:appedndix1} shows the language scores computed between the two types of descriptions from the original AVSD dataset~\cite{alamri2019audio,hori2019end} using the ground truth captions obtained after watching the entire video (\textit{i.e.}, the input descriptions) as references.
The average word lengths of the input and final descriptions are 23.8 and 23.0 words, respectively. Although the word lengths are rather similar, the relatively low scores between two types of descriptions reveal their evident disparity.

\begin{table}[h]
    \center
    \caption{Language scores between the input and final descriptions from the AVSD dataset}
    \scalebox{0.92}{
    \label{tab4:appedndix1}
    \begin{tabular}{|c|c|c|c|c|c|}
    \hline
    Metric     & BLEU4        & METEOR  & SPICE       & ROUGE\_L      & CIDEr    \\   \hline
    GT annotations    & 5.12 & 15.0 & 16.9 & 27.0 & 26.5 \\ \hline
    \end{tabular}}
\end{table}

Figure~\ref{fig:sup1} shows more qualitative examples from the AVSD dataset that reveals the difference between the input and the final descriptions as well as the knowledge gap. We also provide the raw frames and the segmented frames for comparisons. We observe from the figure that the final descriptions given by human annotators lack certain details compared to the input descriptions, which are the video captions obtained after watching the entire video. 

%%%%%%%%%%%%%%%%%%%%%%%%%%%%%%%%%%%%%

\section{Details about the Model}

We list the details about the model components used in our experiments in this section in Table~\ref{table5:appendix2}. In addition, for the LSTM layers, we use the Xavier weight initialization~\cite{jia2014caffe}.

\begin{figure*}[th]
    \centering
    \includegraphics[width=0.95\textwidth]{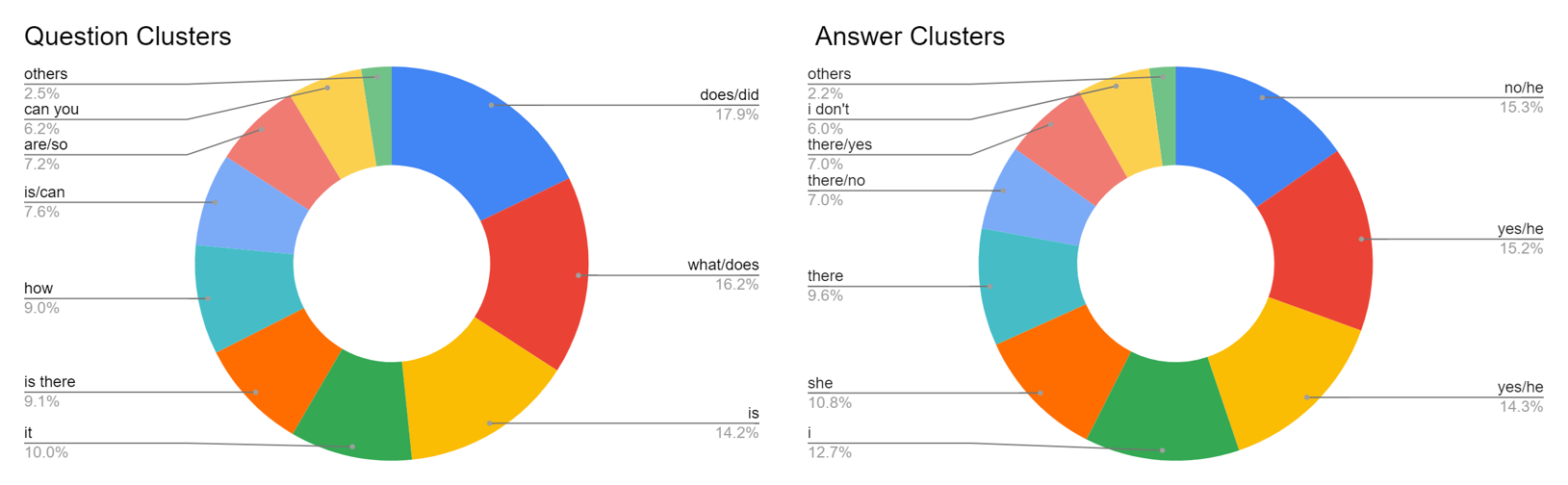}
    \caption{Distributions of the clusters for question and answer candidates in inference. We roughly show the first n-grams for the majority of questions and answers in each cluster. It is worth noting that there are possibly several clusters with similar first n-grams due to the fact that we embed the entire sentence for clustering.}
    \label{fig:sup2}
\end{figure*}

\begin{figure*}[th]
    \centering
    \includegraphics[width=0.95\textwidth]{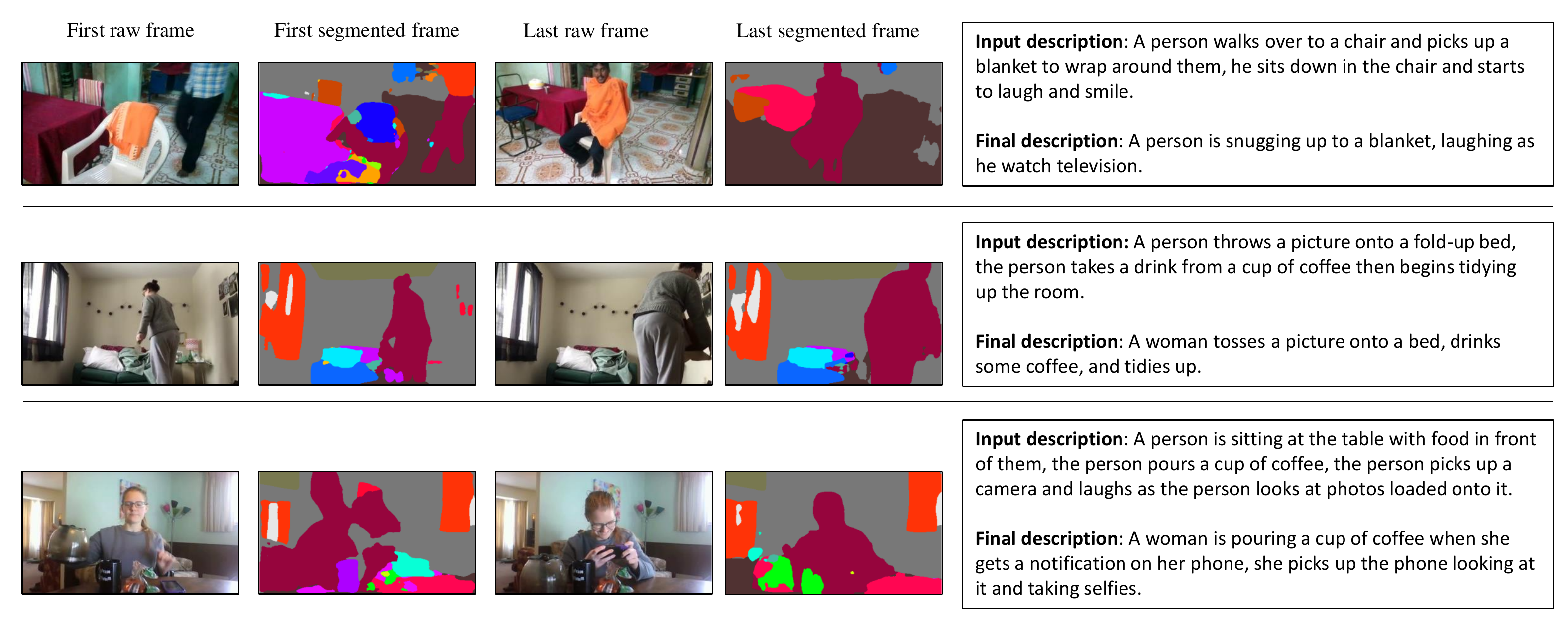}
    \caption{More qualitative examples from the AVSD datasets with input and final descriptions. We observe that the final descriptions given by human annotators without seeing the entire videos miss certain details compared to the input descriptions, despite the dialog interactions help to provide more video information that are not revealed in the static frames.}
    \label{fig:sup1}
\end{figure*}

\section{More evaluation about the internal selection mechanism}

\begin{figure}[t]
    \centering
    \includegraphics[width=0.48\textwidth]{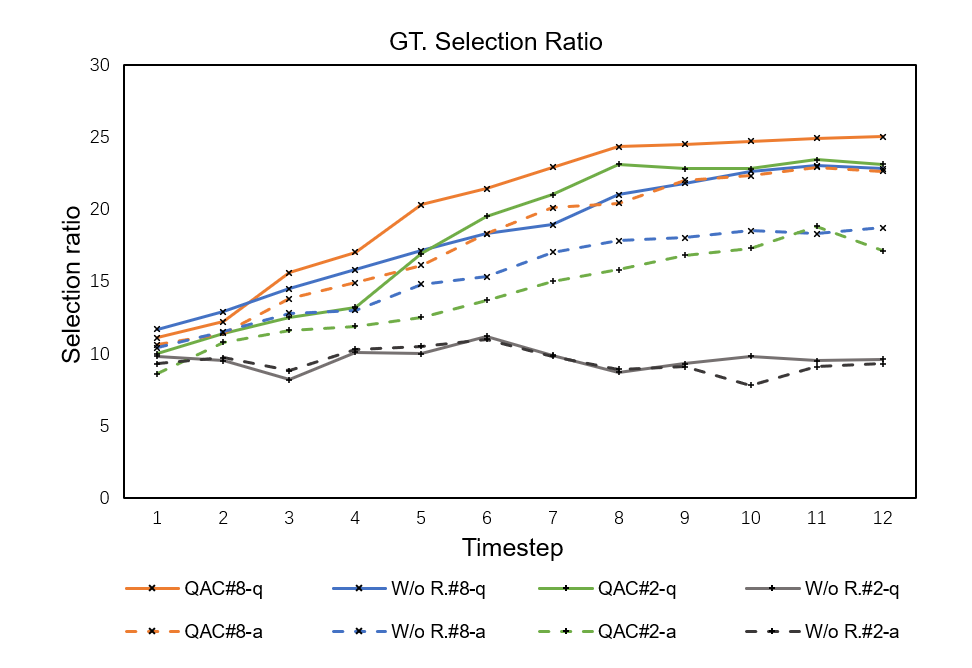}
    \caption{Ground truth question and answer selection ratio during training. We plot the selection ratios with the starting round number 2 and 8 as examples. The solid lines represent the ground truth selection ratios for questions, the dotted lines are the ratios for answers.}
    \label{fig3:reasoning}
\end{figure}

In addition to the final description performance in \emph{w/o Reasoning} setting, we also calculate the ground truth question and answer selection ratio as qualitative evaluation as shown in Figure~\ref{fig3:reasoning}. 
The ground truth selection ratios increase after deploying the internal reasoning mechanism. Additionally, we also observe that the selection ratio for questions is generally higher than the ratio for answers.
The selection ratios also increase as more ground truth QA pairs are provided as input (\textit{i.e.}, with larger starting round number), as in Figure~\ref{fig3:reasoning}, the ground truth selection ratios with the starting round number 8 are generally higher than the starting round number 2.

%%%%%%%%%%%%%%%%%%%%%%%%%%%%%%%%%%
\section{More qualitative results}

We provide more qualitative results and analysis in this section. 

\begin{figure*}[t]
    \centering
    \includegraphics[width=0.99\textwidth]{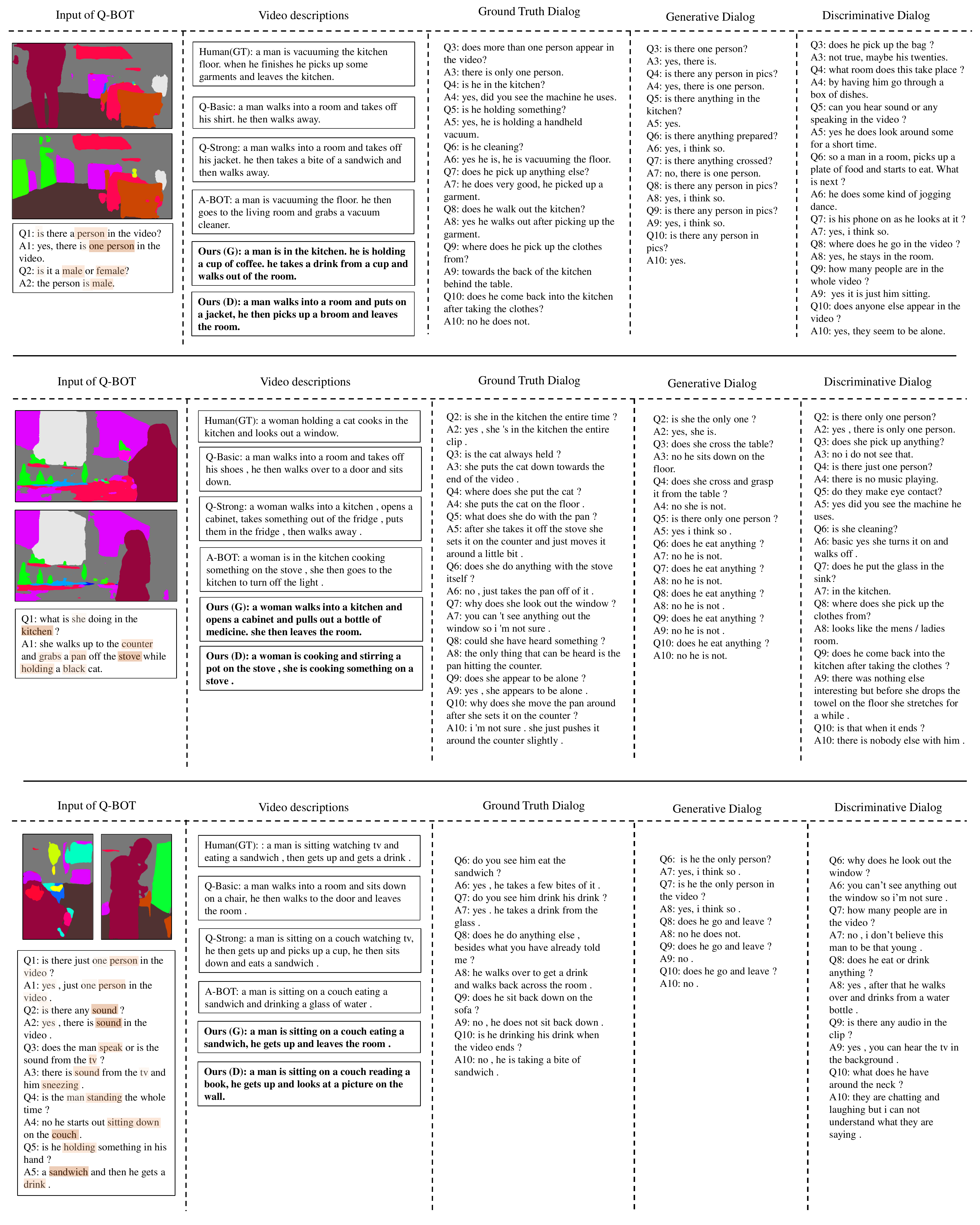}
    \caption{More qualitative results for the proposed task in addition to the example shown in the main paper.}
    \label{fig:quali}
\end{figure*}

\noindent
\textbf{Additional Qualitative Examples.}
We present in Figure~\ref{fig:quali} more qualitative examples. 

\noindent \textbf{Question Evaluation.}
Although the final objective of our work is for \emph{Q-BOT} to describe an unseen video, the ability of \emph{Q-BOT} to ask meaningful questions is also very important. 

For the generative setting, there is no explicit loss function for the question generation process imposed on the \emph{Q-BOT} during the training, therefore, the model tends to ask repetitive questions with a relatively high score of Self-BLUE4 metric~\cite{zhu2018texygen} of 0.82. We then proceed to introduce the discriminative setting to reduce the possible bias learned from the generative setting.

Figure~\ref{fig:sup2} shows the distributions of the clusters for question and answer candidates in inference for the discriminative dialog setting.

\noindent \textbf{Human Evaluation.}
Considering that the intended practical application scenario for our proposed task involves the interactions between the AI systems (\textit{i.e.}, Q-BOT) and real human users (\textit{i.e.}, A-BOT), we perform an extra set of human evaluation test to provide a more thorough analysis of our work. 

\begin{figure*}[t]
    \centering
    \includegraphics[width=0.95\textwidth]{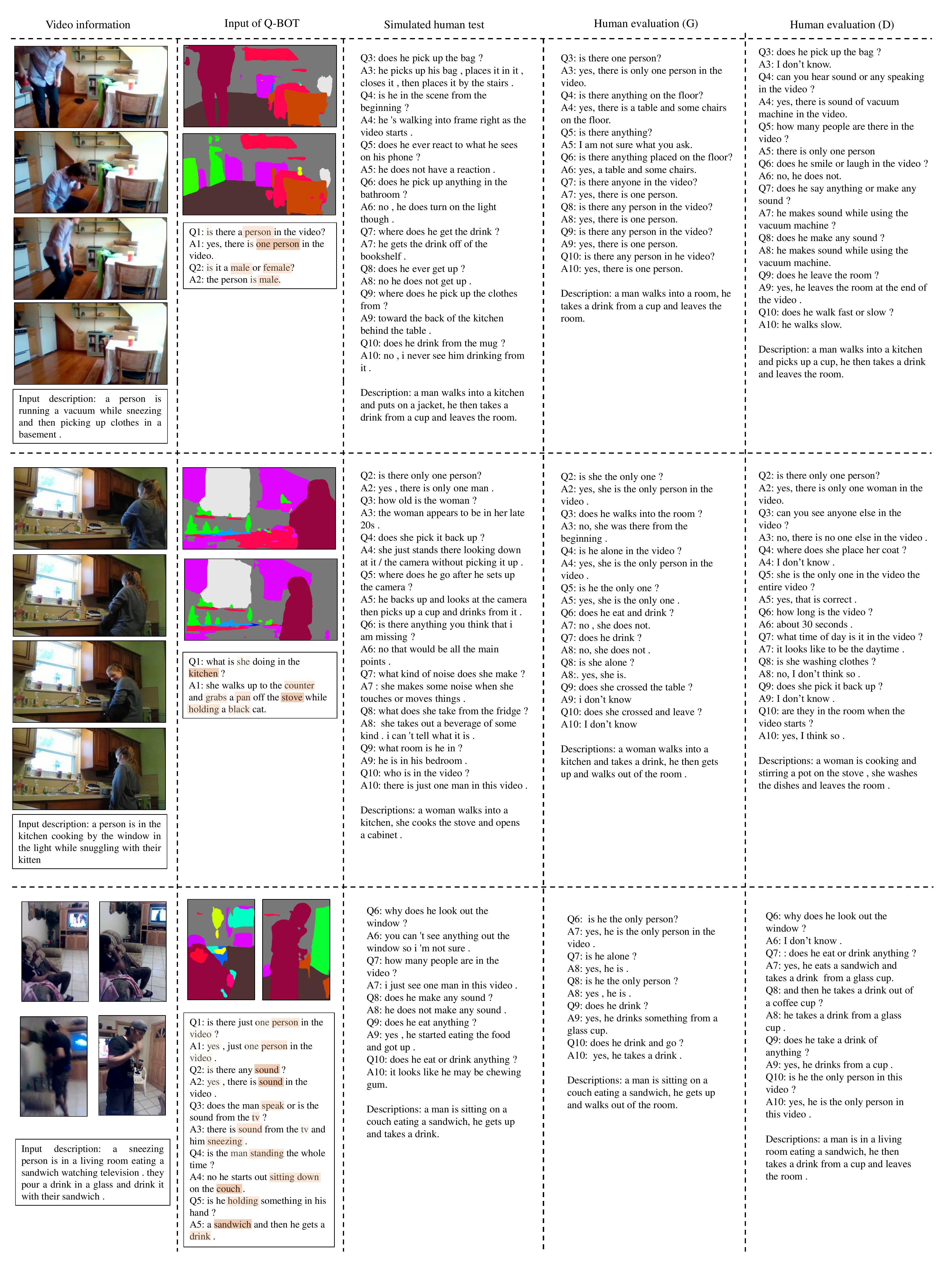}
    \caption{More qualitative results for the simulated human test and real human evaluations.}
    \label{fig:sup3}
\end{figure*}

Figure~\ref{fig:sup3} shows qualitative examples of the human evaluation test corresponding to the qualitative examples in the main paper. 
During the test, we replace the role of \emph{A-BOT} by human participants and provide the real-time answers according to the generated/selected questions. It is worth mentioning that there are several difficulties during the human evaluation test. The most challenging problem is that the questions asked by \emph{Q-BOT} are not always reasonable. Specifically, there are questions that are irrelevant to the actual video. We intentionally define that for those questions, the participants can always provide the answer as "I don't know".  

\begin{table}[h]
\centering
\caption{Details about the model components. The column of agent without specification of \textbf{G}enerative or \textbf{D}iscriminative means the component is the same for both settings.}
\label{table5:appendix2}
\scalebox{1.0}{
\begin{tabular}{|m{0.8cm}<{\centering}|m{1.5cm}<{\centering}|m{2cm}<{\centering}|m{2.6cm}<{\centering}|}
\hline
Agent & Component & Functions & Details \\ \hline
Q\&A & MM module & cross-modal attention & \cite{schwartz2019factor} \\ \hline
Q\&A & history encoder & process the existing dialog history & linear + 1-layer LSTM with size equals to the dimension of the history embedding 256 \\ \hline
Q & visual LSTM & process the segmented visual input & LSTM with 2 units with size equals to the dimension of the attended visual embedding 128 \\ \hline  
Q(G) & question decoder & generate questions to ask & LSTM-based generator wit size equals to the dimension of the question 128 \\ \hline
Q(D) & question decoder & select questions to ask & linear + dot product + softmax selection \\ \hline
Q(D) & candidates encoder & process the question candidates & 1-layer LSTM with size equals to the dimension of question embedding 128 \\ \hline
Q & descrition generator & generate the final video descriptions & LSTM-based generator with size equals to the overall history embedding 256 \\ \hline
A & audio-visual LSTM & process the audio and video information & LSTM with 5 units with 
size equals to the dimension of attended audio + visual embedding 256 \\ \hline
A & input description encoder & process the input description & linear + 1-layer LSTM with size equals to the dimension of the input description embedding 256 \\ \hline
A(G) & answer decoder & generate answers for the questions raised by Q-BOT & LSTM-based generator with size equals to the dimension of question embedding + history embedding + input description embedding \\ \hline
A(D) & answer decoder & select answers for the questions picked by Q-BOT & linear + dot product + softmax \\ \hline
A(D) & candidate encoder & process the answer candidates & 1-layer LSTM with size equals to the dimension of answer embedding 128 \\ \hline
Q\&A & dynamic history update & update the existing history and emphasize the newly generated information & linear + concatenation \\ \hline
\end{tabular}}
\end{table}

\end{document}